\documentclass[10pt,twocolumn,letterpaper]{article}

\usepackage[pagenumbers]{wacv}

\usepackage[T1]{fontenc}
\usepackage{booktabs}
\usepackage{booktabs}
\usepackage{graphicx}
\usepackage{amssymb}
\usepackage{colortbl}
\usepackage{float}
\usepackage{listings}
\usepackage{textcomp}

\definecolor{primarypastel}{RGB}{226,239,252}
\definecolor{resistancepastel}{RGB}{229,245,232}
\definecolor{utilitypastel}{RGB}{253,235,218}
\definecolor{jsonstring}{RGB}{35,110,55}
\definecolor{jsonnumber}{RGB}{145,55,145}
\definecolor{jsonliteral}{RGB}{30,85,170}

\lstdefinelanguage{json}{
    showstringspaces=false,
    breaklines=true,
    string=[s]{"}{"},
    stringstyle=\color{jsonstring},
    literate=
     *{"}{{\textquotedbl}}{1}
      {0}{{{\color{jsonnumber}0}}}{1}
      {1}{{{\color{jsonnumber}1}}}{1}
      {2}{{{\color{jsonnumber}2}}}{1}
      {3}{{{\color{jsonnumber}3}}}{1}
      {4}{{{\color{jsonnumber}4}}}{1}
      {5}{{{\color{jsonnumber}5}}}{1}
      {6}{{{\color{jsonnumber}6}}}{1}
      {7}{{{\color{jsonnumber}7}}}{1}
      {8}{{{\color{jsonnumber}8}}}{1}
      {9}{{{\color{jsonnumber}9}}}{1}
      {true}{{{\color{jsonliteral}true}}}{4}
      {false}{{{\color{jsonliteral}false}}}{5}
      {null}{{{\color{jsonliteral}null}}}{4}
}

\lstdefinestyle{json}{
    language=json,
    basicstyle=\ttfamily\scriptsize,
    columns=fullflexible,
    keepspaces=true,
    upquote=true,
    breakatwhitespace=false,
    frame=single,
    framerule=0.3pt,
    rulecolor=\color{black!25},
    backgroundcolor=\color{black!2},
    xleftmargin=2pt,
    xrightmargin=2pt,
    aboveskip=3pt,
    belowskip=3pt
}

\newcommand{\ssecref}[1]{Section~\ref{ssec:#1}}
\newcommand{\secref}[1]{Section~\ref{sec:#1}}
\newcommand{\figref}[1]{Figure~\ref{fig:#1}}

\definecolor{wacvblue}{rgb}{0.21,0.49,0.74}
\usepackage[pagebackref,breaklinks,colorlinks,allcolors=wacvblue]{hyperref}

\newcommand{\ieeenotice}{\begingroup
  \renewcommand{\thefootnote}{}%
  \footnotetext{This work has been submitted to the IEEE for possible
  publication. Copyright may be transferred without notice, after which this
  version may no longer be accessible.}%
  \endgroup}

\title{CellTrack-Bench: Microscopic Cell Tracking \\ Benchmark for Vision--Language Models}

\author{%
  Ilan Theodoro \quad Jord\~ao Bragantini \quad Lo\"ic A.~Royer \\
  Biohub \\
  San Francisco, CA \\
  \texttt{\{ilan.silva, jordao.bragantini, loic.royer\}@biohub.org}
}

\begin{document}
\maketitle
\ieeenotice
\begin{abstract}
Large language models have shown outstanding performance on text-first tasks  such as coding, math, and writing. Recent efforts are shifting models' training  to include scientific problems that often live in highly specialized domains  beyond text, such as protein structure and biological images. However, in these  scenarios, their general-purpose accuracy compared with specialized algorithms  remains in question. One such domain is cell tracking, which requires  spatiotemporal awareness of cell dynamics. To evaluate this, we created a new  benchmark, CellTrack-Bench, which assesses Vision-Language Models' (VLMs)  cell-tracking-related performance against specialized trackers. The full  benchmark draws from 15 standard cell-tracking datasets across four imaging  modalities and comprises 1,537 problems corresponding to 9,041 cell-linking decisions. We selected a hard subset requiring greater reasoning effort by identifying failures shared across distinct algorithmic trackers. With this proposed dataset, we: (i)  evaluated different environment harnesses for VLMs; (ii) benchmarked 9 frontier  closed-source models and 13 open-weight models; and (iii) provided a detailed  analysis of their failure modes. Code, the stored model outputs, and a pipeline that
regenerates the benchmark from the source datasets are available at \url{https://github.com/royerlab/celltrack-bench}.
\end{abstract}
    
\section{Introduction}

Cells exhibit complex behaviors, including proliferation, differentiation, and migration, throughout their life, resulting in complex biological patterns such as embryonic
development~\cite{lange2024multimodal, mcdole2018toto}, tissue mechanics~\cite{ichbiah2023embryo}, and immune response~\cite{laviron2019tracking}. To 
quantify this dynamic cellular behavior at the single-cell level, it is 
necessary to identify individual cells and to track them over 
time~\cite{meijering2009tracking}.

Cell tracking shares many similarities with Multi-Object Tracking (MOT)~\cite{dendorfer2020mot20} in the traditional natural imaging domain, as both involve identifying and tracking objects.
They're commonly solved through the paradigm of tracking-by-detection, where an algorithm segments or detects the objects and then links them over time.
However, cell tracking benefits from specialized
solutions~\cite{ulman2017objective,mavska2023cell, bragantini2025ultrack} because
objects are almost identical, can be divided, and quickly undergo morphological 
changes, unlike objects in natural images (e.g., cars, pedestrians).

\begin{figure}[t]
  \centering
  \includegraphics[width=\columnwidth]{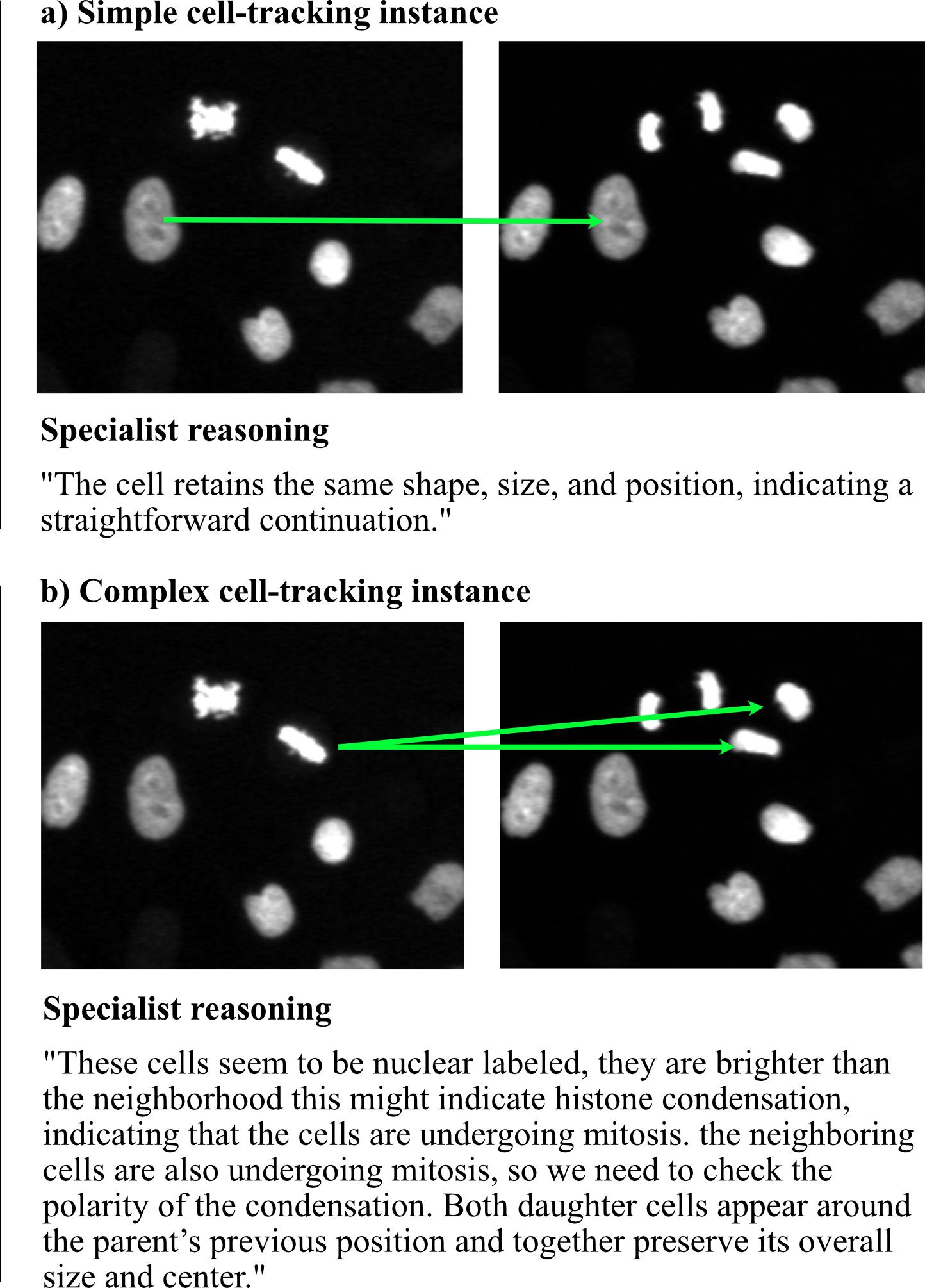}
  \caption{%
  \textbf{Cognitive aspect of cell tracking.} Cell tracking is dominated by a majority of trivial associations that can be retrieved using simple proxies such as a proximity-based one. In \textbf{(a)}, we show a human specialist cognitive process on one of those links. However, in some cases of cell-state changes, such as a division in \textbf{(b)}, it requires a more global comprehension and understanding of biological processes.}
  \label{fig:cognitive_reasoning}
\end{figure}

Here, we take a different approach. Rather than dividing the problem from the algorithmic point of view, we view the operations in two different regimes: (i) a highly mechanical scenario 
where it's easy to design an algorithm to automatically link the objects, for 
example, with optimal matching~\cite{jaqaman2008robust, fukai2023laptrack}; and 
(ii) a highly cognitive scenario where one must reason about what is happening 
within the image, considering cell state, if undergoing division, if near the 
boundary of the field of view, and many other experimental conditions, see Figure~\ref{fig:cognitive_reasoning} for a practical example.

This motivates the development of a system where existing cell tracking algorithms perform 
the mechanical aspect of tracking, while VLMs perform the highly cognitive 
correction of mistakes made by these approaches. This is analogous to a 
scenario where an annotator is fixing the tracking mistakes. 
To validate this novel paradigm for cell tracking, we built CellTrack-Bench from
15 cell-tracking datasets across four imaging modalities. Thirteen datasets
contribute 1,537 problems, amounting to 9,041 linking instances.

For CellTrack-Bench, we define three tasks that progressively change the role of the VLM from directly constructing associations, to correcting an existing tracking solution, to identifying errors for review: (i) cell linking; (ii) track repair; and (iii) track inspection. 

In \textbf{cell linking}, the model directly predicts the correspondence between cells in two consecutive frames. In \textbf{track repair}, it is given a graph from an algorithmic tracker and has to determine the set of link corrections that will fix the graph. In \textbf{track inspection}, it detects the tracker decisions that need review, corresponding to a human-in-the-loop triage scenario.  Taken together, these tasks highlight different aspects of tracking that the model can collectively cover the operations needed to correct tracking errors across a dataset.

\begin{figure*}[t]
  \centering
  \includegraphics[width=\textwidth]{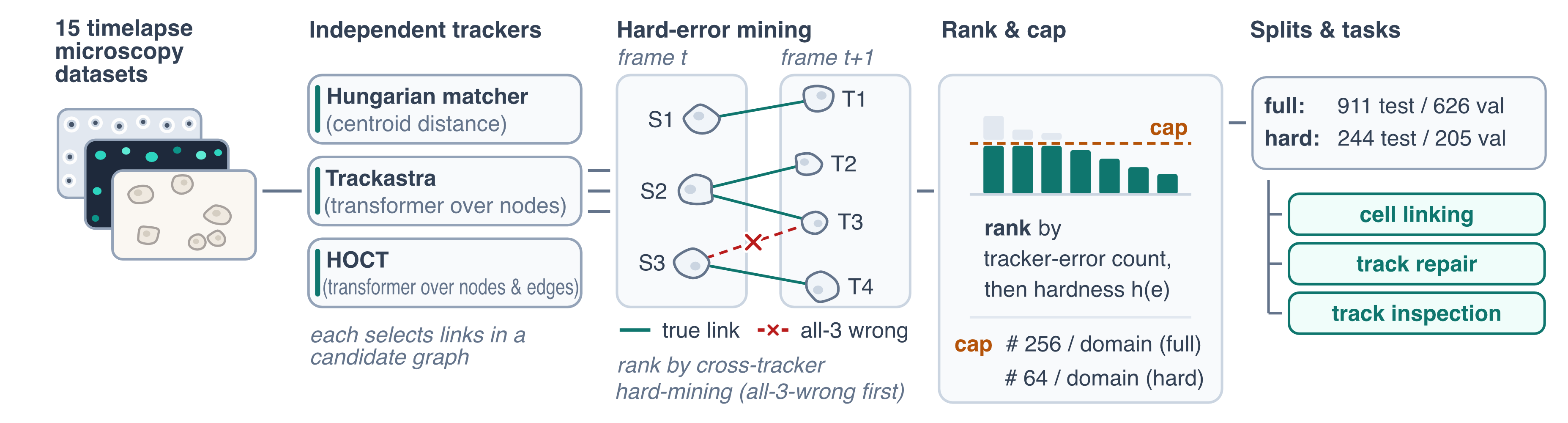}
  \caption{%
  \textbf{CellTrack-Bench construction.} Three trackers with distinct inductive biases run across four imaging modalities; errors are ranked by a cross-tracker hard-mining score, all-three-wrong first (illustrated: the division $S2 \rightarrow \{T2, T3\}$ that every tracker resolves as $S3 \rightarrow T3$), then by hardness $h(e)$, backfilled in rank order to a per-dataset cap, split sequence-disjointly, and instantiated as linking, repair, and inspection tasks (scales and counts: \ssecref{source_data}).
  }
  \label{fig:pipeline}
\end{figure*}

We apply these three tasks to difficult cell tracking cases that are local rather than to complete time-lapse sequences. In a normal cell-tracking problem, an algorithm handles a complete time-lapse series of microscopy images; in our approach, each problem is concentrated on the more difficult cell links (see \figref{pipeline}). This is necessary 
since, in most datasets having sufficient temporal resolution, cells do not 
move much until a cell-state change takes place, at which point their visual 
appearance may change and they can exhibit erratic movement and displacement. 
Although such events are rare and more prone to error, the overall tracking 
performance can mask them. We identify the difficult cases algorithmically by 
looking at failures that are common to multiple trackers.

After having chosen the difficult cases and defined the tasks, assessing the VLMs also required controlling the way in which these problems are presented to the model. For this reason, we conducted a study looking at various environment strategies, recently known as \emph{harnesses}, such as the method of presenting cells, the way in which coordinate information is given, the use of multi-turn settings, and visual hints like segmentation outlines.
For our 
study we carried out a validation-only investigation into environmental changes 
with respect to the VLM and kept the best-performing configuration for the test 
evaluation. We also introduce proximity-debiased metrics which separately 
assess associations that are not already accounted for by a nearest-neighbor 
rule, together with division-specific and graph-level measures. Our protocol 
also takes into account the composition of the dataset
and the output-token budget.

In summary, our contributions are:

\begin{itemize}
    \item A benchmark comprising 1,537 hard-mined adjacent-frame problems,
with 9,041 linking associations spanning 15 datasets and four imaging modalities, 
structured as a series of tasks concerning cell linking, track repair, and 
track inspection.

    \item A controlled evaluation protocol for VLM-based cell tracking, 
including validation-only harness selection, proximity-debiased, and 
division-specific metrics, a Cell Tracking Challenge (CTC) comparable graph 
score, dataset-stratified evaluation, and inference 
budget.

    \item A study evaluating 9 frontier closed-source and 13 open-weight VLMs by means
of a fixed interface, together with geometric and specialized tracking 
references, looking at proximity shortcuts, interface sensitivity, and token budget 
effects, and task relationships.

\end{itemize}

This work is organized as follows. The \secref{related_works} positions our 
work within the literature. In \secref{methods}, we explain how the benchmark 
is created and how we define the evaluation protocol. Finally, in 
\secref{results} we conduct a discussion over the results of closed-source and 
open-weight models, followed by our final considerations in \secref{conclusion}.

\newcommand{\dscell}[2]{%
  \begin{subfigure}[t]{0.31\columnwidth}
    \centering
    \includegraphics[width=\linewidth]{figures/datasets/#1}%
    \caption*{\scriptsize #2}
  \end{subfigure}}

\begin{figure}[t]
  \centering

  \dscell{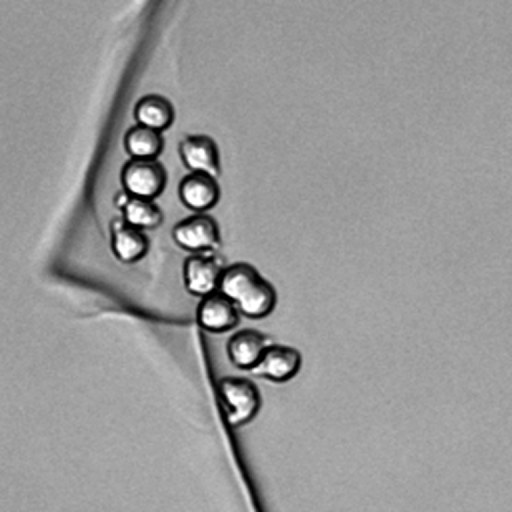}{BF-C2DL-HSC}\hfill
  \dscell{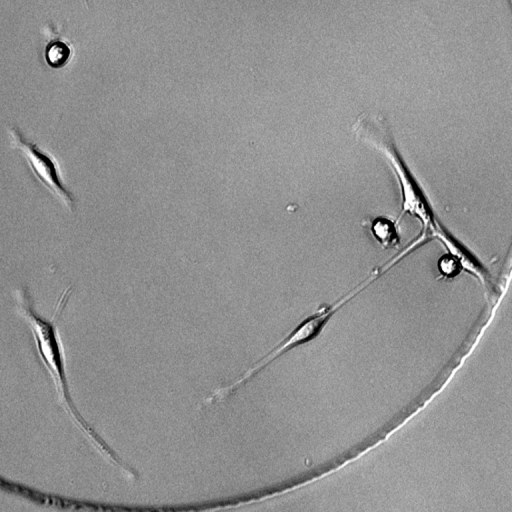}{BF-C2DL-MuSC}\hfill
  \dscell{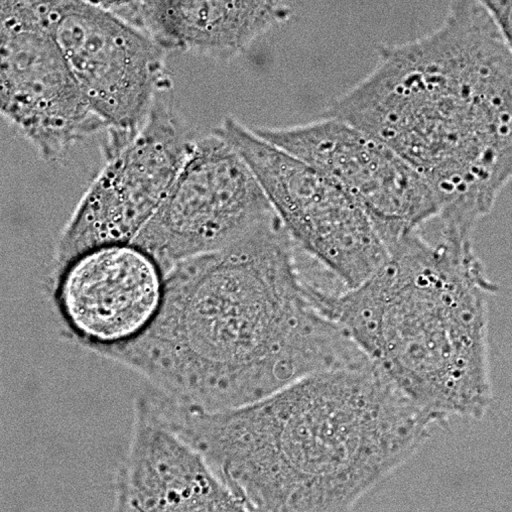}{DIC-C2DH-HeLa}\\[4pt]

  \dscell{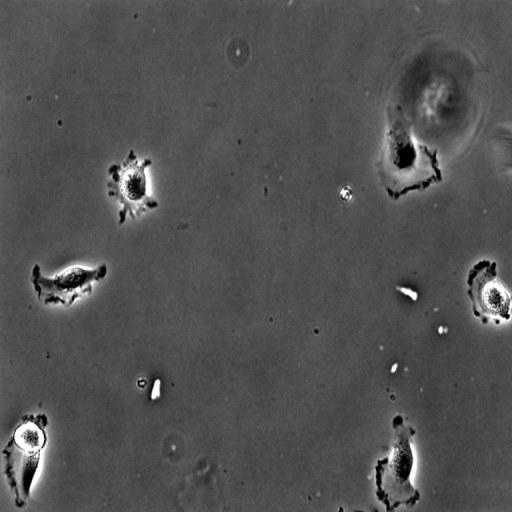}{PhC-C2DH-U373}\hfill
  \dscell{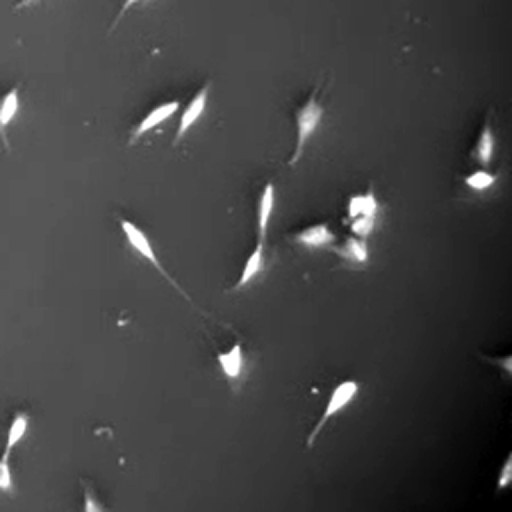}{PhC-C2DL-PSC}\hfill
  \dscell{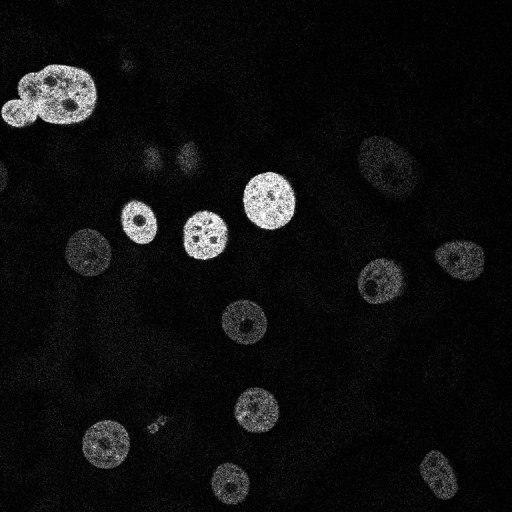}{Fluo-N2DH-GOWT1}\\[4pt]

  \dscell{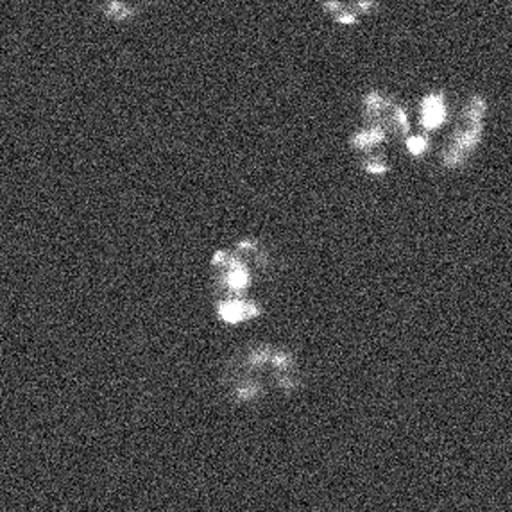}{Fluo-N2DH-SIM+}\hfill
  \dscell{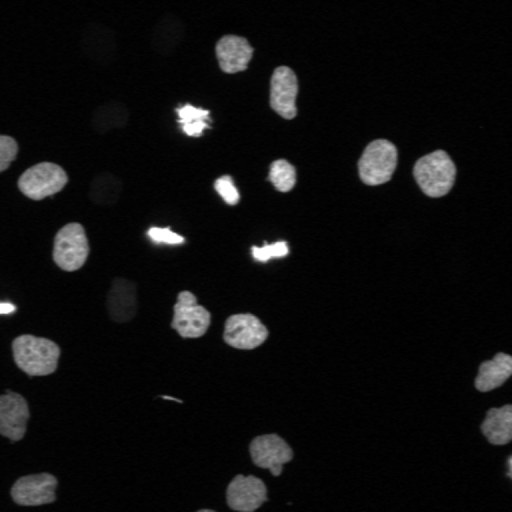}{Fluo-N2DL-HeLa}\hfill
  \dscell{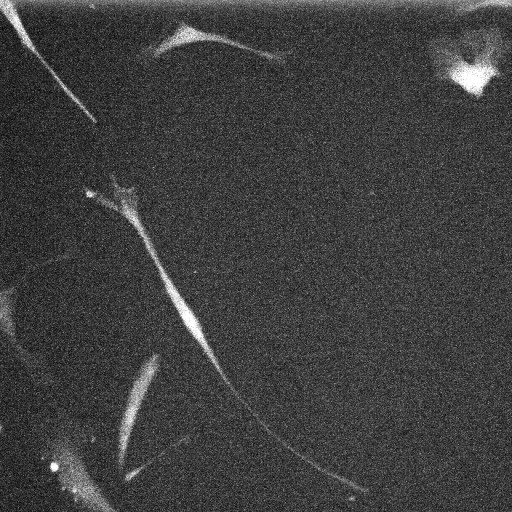}{Fluo-C2DL-MSC$^{*}$}\\[4pt]

  \dscell{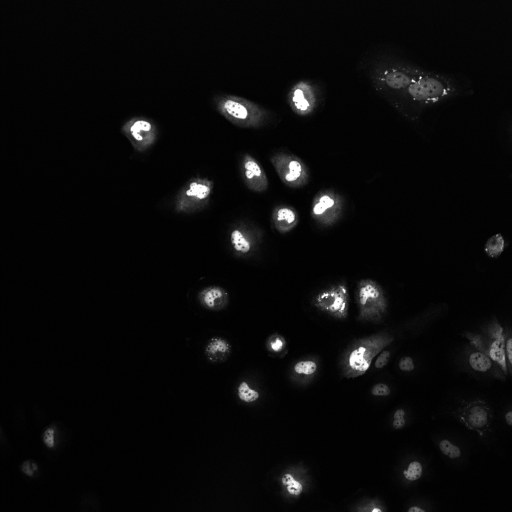}{Fluo-C2DL-Huh7$^{*}$}\hfill
  \dscell{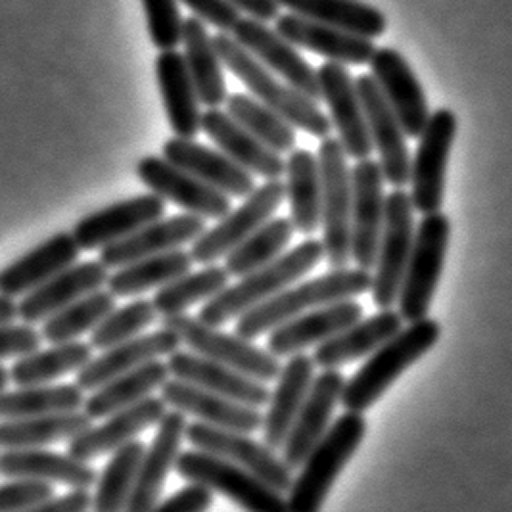}{Bacteria-cib}\hfill
  \dscell{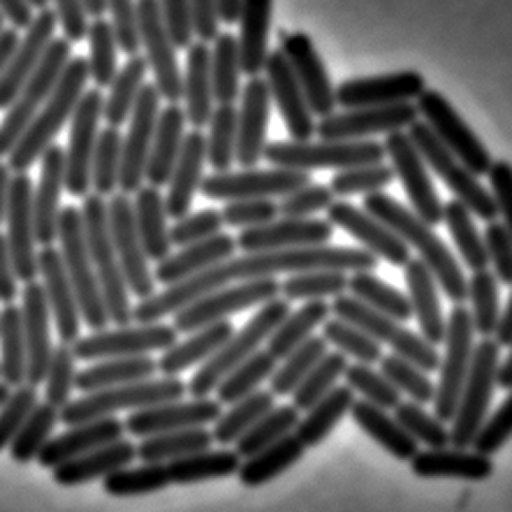}{Bacteria-metA}\\[4pt]

  \dscell{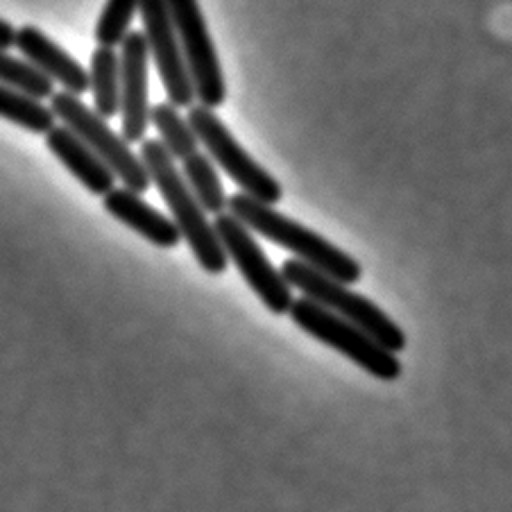}{Bacteria-pheA}\hfill
  \dscell{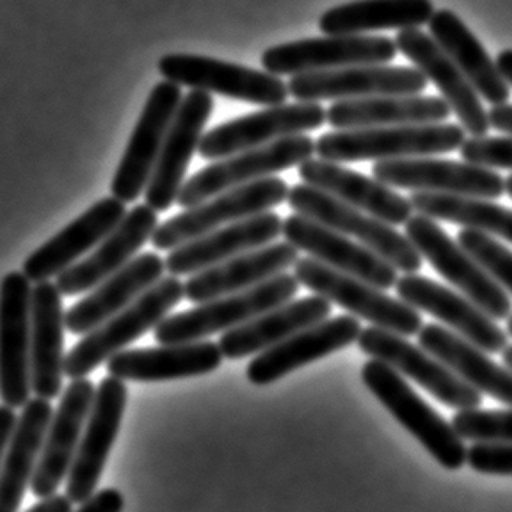}{Bacteria-rpsM}\hfill
  \dscell{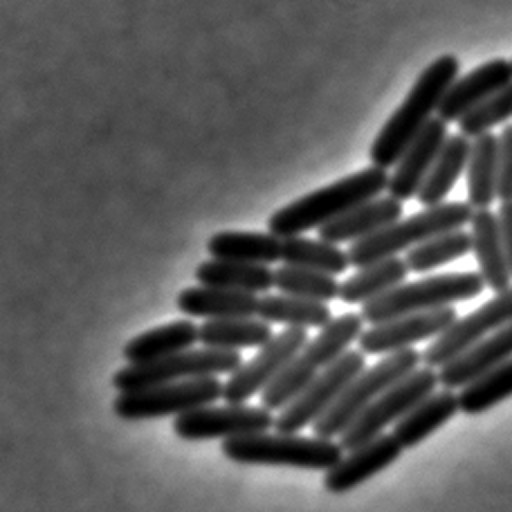}{Bacteria-trpL}%

  \caption{Example frames from all 15 source datasets, spanning four imaging modalities:
  bright-field (BF), differential interference contrast (DIC), phase contrast (PhC),
  and fluorescence (Fluo).}
  \label{fig:dataset_gallery}
\end{figure}

\section{Related Works}
\label{sec:related_works}

\paragraph{Cell-tracking methods and benchmarks.}

With recent advancements in instance segmentation 
models~\cite{stringer2021cellpose,goldsborough2024instanseg,schmidt2018cell}, 
cell-tracking methods have almost exclusively relied on the 
tracking-by-detection paradigm, focusing on association (edge) prediction, as 
TrackMate~\cite{jaqaman2008robust,tinevez2017trackmate} and 
btrack~\cite{ulicna2021automated} on the algorithmic side. On the deep learning 
front, Trackastra~\cite{gallusser2024trackastra} predicts pairwise cell 
detection associations using a transformer over temporal windows, while 
HOCT~\cite{bragantini2026higher} models the higher-order cell-to-cell 
interactions among the competing candidate links. The most widely regarded 
benchmark is the Cell Tracking Challenge (CTC), which standardized performance 
evaluation using lineage-aware graph-edit 
metrics~\cite{ulman2017objective,mavska2023cell} and includes a hidden test 
evaluated by the organizers. Our proposed benchmark uses a similar methodology, 
assuming cell segmentation is given, and evaluating performance based on local 
associations between consecutive frames. 

Because even very effective trackers still produce a few mistakes,
human-in-the-loop review is often necessary. Pop et al.~\cite{pop2025needles} framed
the verification as a multi-armed bandit problem. OrganoidTracker flags 
suspicious links for review~\cite{kok2020organoidtracker}. Meanwhile, its 
successor, OrganoidTracker 2, computes probabilities for each 
link~\cite{betjes2025cell}. These interactive approaches motivate our 
track repair and track inspection tasks, in which a VLM serves as an external
assistant that either corrects or prioritizes links for revision. Because the 
difficulty of tracking individual cells is heavily skewed towards easy cells, 
existing metrics fail to assess the quality of hard instances; therefore, 
CellTrack-Bench extends evaluation from tracker-specific metrics to a more 
specific assessment of the model's ability to link, repair, and inspect hard 
instances across multiple trackers and imaging modalities.

\textbf{Biomedical VLM benchmarks.}
The $\mu$-Bench~\cite{lozano2024micro} attempts to measure perception and 
cognition with respect to light, fluorescence, and electron microscopy. 
MicroVQA~\cite{burgess2025microvqa} focuses on expert scientific reasoning 
using single- and multi-image questions and answers. 
OmniMedVQA~\cite{hu2024omnimedvqa} evaluates medical image understanding on a 
large scale. Both MedFrameQA~\cite{yu2025medframeqa} and general video 
benchmarks like MVBench~\cite{li2024mvbench} also look into temporal reasoning. 
These benchmarks primarily consist of classification, localization, or 
question-answering tasks. In contrast, CellTrack-Bench evaluates cell-identity 
correspondence, including continuations, divisions, appearances, and 
disappearances, as well as explicit changes to lineage graphs.

\textbf{Shortcut-resistant and controlled evaluation.}
Benchmark performance might overestimate a model's visual understanding if the 
model makes use of linguistic or visual 
shortcuts~\cite{geirhos2020shortcut,chen2024we}. In cell tracking, the main 
shortcut is spatial proximity: a nearest-neighbor rule accounts for much of the 
association distribution. CellTrack-Bench therefore uses sequence-disjoint 
validation/test splits, implements a ranked tracker-error mining, per-dataset
sampling caps, and uses debiased metrics with respect to proximity or division 
events. VLM performance can also depend on how the problem is presented and the 
models' harness~\cite{schema2026}. Therefore, the evaluation protocol includes 
the renderer, the coordinate exploration interface, the prompt, and the 
inference budget. We selected these components based on validation data and 
kept them fixed during testing. The result is a controlled diagnostic benchmark 
for VLM-assisted cell linking and track inspection, not a new tracker or training
method.

\section{CellTrack-Bench}
\label{sec:methods}

In this section, we explain how we constructed CellTrack-Bench. First, in 
\ssecref{benchmark_setting} we describe the standard protocol used in the Cell 
Tracking Challenge (CTC)~\cite{mavska2023cell,ulman2017objective}, and in 
\ssecref{source_data} we document the data we gathered. Then, in 
\ssecref{difficulty_probes}, we explain how we use state-of-the-art and 
classical tracking algorithms to identify hard instances that may indicate data 
requiring stronger cognitive abilities, followed 
by how we transformed these instances into real tasks for the VLMs in \ssecref
{benchmark_tasks}. 
In \ssecref{metrics_and_evaluation}, we then propose the metrics and evaluation 
protocol used throughout this work.
Finally, we empirically explore prompting and harnessing design choices to 
enable models to find solutions more effectively in 
\ssecref{prompting_and_harnessing}.

\subsection{Preliminaries}
\label{ssec:benchmark_setting}

The tracking problem is formulated as constructing a binary forest (i.e., each 
node can have at most two children) from a candidate graph, $G_C = (V_C, E_C)$, 
which contains a superset of valid tracking solutions; throughout the text, we 
will refer to the predicted graph as a solution graph $G_{pred} = (V_{pred}, 
E_{pred})$ and the ground-truth graph computed from the annotations as $G_{gt} 
= (V_{gt}, E_{gt})$.

The nodes $V_{\cdot}$ from all graphs come from each frame $t$ of the 
segmentation sequence (video), such that segments from a single frame are $V^t 
\subset V_C$.

CellTrack-Bench retains the images, segmentations, and lineage annotations from 
existing tracking datasets but changes the unit of evaluation from a complete 
sequence to local transitions, $e_{ij} \in E_{gt} \cup E_{pred}$ between frames 
$t$ and $t+1$ where $i \in V^t$ and $j \in V^{t+1}$, allowing errors to be 
isolated and diagnosed at the frame level. Each instance contains a compact 
group of source cells that includes the specific error, and the linking, 
repair, and inspection tasks share the same local transition (construction 
detail in \ssecref{benchmark_tasks}).

\subsection{Source Data}
\label{ssec:source_data}

CellTrack-Bench draws from 10 CTC and five bacterial datasets (35 sequences),
encompassing fluorescence, phase-contrast, differential-interference-contrast,
and bright-field microscopy. Object size, density, morphology, and
motion vary widely. See Figure~\ref{fig:dataset_gallery}.

We partition sequences before ranking or selecting links to avoid temporal 
leakage, so visually adjacent 
transitions from the same lineage cannot appear in both partitions. In CTC 
datasets, we assign sequence \texttt{01} to validation and \texttt{02} to test. 
Similarly, we split the multi-sequence bacterial datasets into 
identifier-ordered halves (per-dataset assignments in 
Appendix~\ref{app:benchmark-construction}).
Every benchmark decision is evaluated using the validation dataset, and the 
test dataset was only used for 
the final leaderboard score.

\subsection{Specialized Trackers as Difficulty Probes}
\label{ssec:difficulty_probes}

We argue that trackers' errors are not independent and likely identify mistakes 
in challenging links. We thus use three different trackers with different 
inductive biases and use their shared errors as a hard-mining signal to find 
such difficult
links and mitigate uncorrelated wrong links, which might be just a tracker's
shortcoming. 

All three trackers jointly fail at $0.28\%$ of the 1.5M candidate links (4,222 links
exactly), so the dependence argument is supported empirically, as we would expect only
$0.0022\%$ shared errors under independence (33.2; stratifying by sequence already
accounts for some sequences being uniformly harder; see
Appendix~\ref{app:shared-error-expectation}). Moreover, given one tracker error, another tracker is
$11.3$--$49.6$ times more likely to commit the same error. The trackers are:
\begin{itemize}
    \item \textbf{Hungarian Matching on Distance}. It is a one-to-one 
assignment baseline based uniquely on centroid distance. As such, it doesn't 
model appearance and divisions. Although simple, this baseline scores fairly 
high on the CTC's \emph{LNK} metric of $0.888$. This reinforces that cell
position is already a strong tracking feature.
    \item \textbf{Trackastra}. This tracking method feeds cell-segment features 
as 
tokens in a transformer network and directly learns the association matrix. Its 
\emph{LNK} score is $0.963$. This difference relative to the linear-assignment 
baseline directly shows its ability to predict over a broader range of
difficulty. 
    \item \textbf{HOCT}. As a natural extension to Trackastra, this is a 
higher-order method that segments the cells and the edges as tokens, it scores 
$0.946$ on \emph{LNK}.
\end{itemize}

Each model is trained and evaluated with k-folds, by splitting the 15 tracking 
datasets into three groups of five datasets each. To get the errors for one 
fold, we train on the 
other two and evaluate on the held-out fold. All trackers are trained using the 
provided segmentation masks, so we focus on association rather than detection 
quality. Appendix~\ref{app:benchmark-construction} provides the tracker roles
and held-out protocol. In the next subsection, we detail how we select the
errors.

\subsection{Tracking Errors to Benchmark Tasks}
\label{ssec:benchmark_tasks}

We defined an extended set of candidate edges, $E_C^t$, for each transition, 
with $E_C^t \subseteq
(V_C^t\cup\{a^t\})\times(V_C^{t+1}\cup\{d^{t+1}\})$, where $a^t$ and $d^{t+1}$
are artificial nodes that represent appearing and disappearing nodes. 
Ground-truth links are $E_{gt}^t\subseteq
E_C^t$ and tracker $m$ produces $E_{pred,m}^{t}\subseteq E_C^t$. An error is a 
missing
ground-truth link, a spurious predicted link, or a link with incorrect 
continuation or division semantics. For a link $e$, let $z_m(e)\in\{0,1\}$ 
indicate that tracker $m$ is mistaken at $e$; the hard-mining count is 
$q(e)=\sum_m
z_m(e)\in\{0, 1,2,3\}$. The two learned trackers also produce link scores 
$s_m(e)$: given binary target $y(e)$, hardness is the mean prediction error
$h(e)=\tfrac{1}{2}\,(|s_{\mathrm{HOCT}}(e)-y(e)|+|s_{\mathrm{trackastra}}(e)-y(e
)|)$. The Hungarian tracker contributes to $q(e)$ but not to $h(e)$ because its 
distance cost is not a calibrated link probability.

For every dataset from~\ssecref{source_data}, we cap top-K scorers in the 
hard-mining metric using function $q$ as the main classifier and $h$ as the 
tie-breaker.
For the harder selection, we select with $K=64$. While for 
the extended benchmark, we select $K=256$. Knowing the locations of the link 
errors, we construct the candidate graph by randomly selecting a compact 
neighborhood of 5 cells in the source frame $t$, provided the source error cell 
is within that neighborhood. Then the target cells are selected based on the 
original candidate cell for each of the 5 source cells. This candidate graph 
will be used in the following format:

\begin{itemize}
    \item \textbf{Cell Linking.} The VLM is informed of all cells in the 
candidate graph and will be instructed to reconstruct the correct linkage. Only 
the links in the hard instances pool are evaluated in this scenario.
    \item \textbf{Track Repair.} The VLM is given one of the solved track 
graph solutions that originated the hard instances. Given its edge set
$E_{pred,m}^t$, it predicts links to add, $A^t$, and remove, $R^t$, which
yield the repaired graph $E_{\mathrm{rep},m}^t=(E_{pred,m}^t\setminus R^t)\cup
A^t$; the objective is whether it can improve the current solution.
    \item \textbf{Track Inspection.} The model does not modify the graph, but
returns the set $F^t$ of asserted or rejected tracker decisions that require 
review. This can be used as a triage system for selecting instances for human
or VLM
annotation.
\end{itemize}

\subsection{Metrics and Evaluation Protocol}
\label{ssec:metrics_and_evaluation}

Let $\mathcal{D}$ be the set of dataset indices and $\mathcal{I}_d$ the 
benchmark
items from dataset $d$, with
$\mathcal{I}=\bigcup_{d\in\mathcal{D}}\mathcal{I}_d$. For task
$\tau\in\{\mathrm{link},\mathrm{repair},\mathrm{inspect}\}$ and item $i$, let
$Y_{i,\tau}$ and $\widehat{Y}_{i,\tau}$ be the sets of ground-truth and
predicted tuples, respectively. These are task edge tuples for linking,
task edges after applying the predicted edits for repair, and complete
review-decision tuples for inspection. Such that
\[
\mathrm{F1}(\widehat{Y},Y)=
\dfrac{2|\widehat{Y}\cap Y|}{|\widehat{Y}|+|Y|}
\]

\textbf{Overall F1}. The primary metric for task $\tau$ is the
dataset-macro average of per-item F1:
\[
\mathrm{Overall\,F1}_{\tau}
=\frac{1}{|\mathcal{D}|}\sum_{d\in\mathcal{D}}
\frac{1}{|\mathcal{I}_d|}\sum_{i\in\mathcal{I}_d}
\mathrm{F1}(\widehat{Y}_{i,\tau},Y_{i,\tau}).
\]
For repair, edit-operation F1 is reported separately.

\textbf{NN-Residual F1}. As we've seen in 
\ssecref{difficulty_probes}, a simple nearest-neighbor policy skews the link 
evaluation into easy instances. Hence, a weak tracker naturally learns to 
link the behaviors of the nearest neighbors but fails at harder instances, 
which might require more cognitive effort. Nearest-neighbor (NN)
matching is therefore a necessary solving policy for a tracker, but
not sufficient. The NN-Residual F1 metric is an attempt to remove the NN bias 
from the evaluation by sampling the F1 score only on links that are not 
explained by such heuristic. Let $\mathcal{U}_i$ be the  task candidate-edge 
tuples for item $i$, $N_i$ the endpoint pairs returned by nearest-centroid 
matching, and $\pi(e)$
the endpoint pair of task edge $e$. 
For
$R_i=\{e\in\mathcal{U}_i:\pi(e)\notin N_i\}$, and  
$\mathcal{R}:=\{i\in\mathcal{I}: R_i\neq\emptyset\}$ where the neighborhoods 
have at least one edge not explained by the nearest-neighbor rule, it follows
\[
\begin{aligned}
\widehat{Y}_i^{\mathrm{NN}}&=\widehat{Y}_{i,\mathrm{link}}\cap R_i,\\
Y_i^{\mathrm{NN}}&=Y_{i,\mathrm{link}}\cap R_i,\\
\mathrm{NN\text{-}Residual\,F1}
&=\frac{1}{|\mathcal{R}|}\sum_{i\in\mathcal{R}}
\mathrm{F1}(\widehat{Y}_i^{\mathrm{NN}},Y_i^{\mathrm{NN}}).
\end{aligned}
\]

\textbf{Division F1}. Similar to the previous item, divisions are infrequent 
and challenging in the distribution of links (see Appendix~\ref {app:data}). 
This infrequency can create a low-recall problem for this class. To diagnose 
this, we create this metric by sampling the F1 score on division links. For
$B_i=\{e\in\mathcal{U}_i:\operatorname{sem}(e)=\mathrm{division}\}$, and 
$\mathcal{B}=\{i\in\mathcal{I}:B_i\neq\emptyset\}$,
\[
\begin{aligned}
\widehat{Y}_i^{\mathrm{div}}&=\widehat{Y}_{i,\mathrm{link}}\cap B_i,\\
Y_i^{\mathrm{div}}&=Y_{i,\mathrm{link}}\cap B_i,\\
\mathrm{Division\,F1}
&=\frac{1}{|\mathcal{B}|}\sum_{i\in\mathcal{B}}
\mathrm{F1}(\widehat{Y}_i^{\mathrm{div}},Y_i^{\mathrm{div}}).
\end{aligned}
\]

\textbf{Subgraph-LNK}. The previous F1 metrics do not use the 
graph-edit metric from the CTC benchmark~\cite{matula2015aogm}, which measures 
the number of operations (e.g., add or remove elements) to convert a predicted 
graph into the ground-truth through a weighted sum. 
We adapt the LNK metric to the disjoint union of the local neighborhoods 
following the original CTC definition. Let $\widehat{E}_{i,\tau}$ be the
edge set induced by task $\tau$ for item $i$, and let 
$N_{\mathrm{miss}}^{i,\tau}$,
$N_{\mathrm{spur}}^{i,\tau}$, and $N_{\mathrm{sem}}^{i,\tau}$ count its
missing edges, spurious edges, and matching endpoint pairs with incorrect edges
semantics relative to $E_{gt,i}$. Define
\[
\begin{aligned}
C_{i,\tau} &= 1.5N_{\mathrm{miss}}^{i,\tau}
              +N_{\mathrm{spur}}^{i,\tau}+N_{\mathrm{sem}}^{i,\tau},\\
C_0 &= 1.5\sum_{i\in\mathcal{I}}|E_{gt,i}|,\\
\mathrm{LNK}_{\mathrm{sub},\tau}
&=1-\frac{\min\{\sum_{i\in\mathcal{I}}C_{i,\tau},C_0\}}{C_0}.
\end{aligned}
\]
Thus, a perfect graph scores $1$ and an empty prediction scores $0$. Each task 
induces a graph for scoring. For linking, this is the predicted graph. For 
repair, it is the proposed graph after applying the predicted edits. For 
inspection, it is the proposed graph after an oracle replaces every flagged
source's outgoing edges with its ground-truth outgoing edges. Because
Subgraph-LNK applies CTC graph-edit costs to local neighborhoods with fixed 
segmentation, it is comparable, but not identical, to full-sequence CTC LNK 
(Appendix~\ref{app:subgraph_lnk}).

\subsection{Prompting and Harness Design}
\label{ssec:prompting_and_harnessing}

As described above, a benchmark instance specifies the task, the candidate 
graph, and the ground-truth graph, and their respective scene from the original 
image data, but not the way in which this information is 
presented to a VLM. As in previous work~\cite{schema2026}, we call this 
\emph{harness} the environment the models are subject to; it includes the image 
rendering, the prompt template, the localization cues, the 
available tools, the generation budget, the answer schema, and the parser. The 
way these elements are chosen can have a significant effect on performance by 
making cell localization and visual comparison easier or more difficult.

We look at eight conditions, as summarized in 
Table~\ref{tab:evidence_harnesses}. Those were chosen considering different 
ways that the model could operate. In each case, we use the same frames, 
candidate graph, task instruction, answer schema, and scorer; the only 
difference is the evidence interface. The conditions vary by whether the cells 
are identified using normalized bounding boxes (text-based) or visual outlines 
(image-based), whether 
close-up crops are provided in advance or requested by the model, and whether 
the global two-frame overview is available. They involve two main types of 
interface: an \emph{evidence-supplied} single-turn interface, which offers 
annotated frames and pre-rendered cell crops, and an \emph{evidence-requested} 
multi-turn interface, in which the model chooses the areas to examine using a 
\texttt{crop\_and\_zoom} tool. Condition c7 omits the global overview and is 
considered merely as an information-loss ablation.
Appendix~\ref{app:model-harness} illustrates the two interface families and
all eight evidence presentations in Figures~\ref{fig:multiturn}
and~\ref{fig:conditions}, respectively.

\begin{table*}[t]
    \centering
    \scriptsize
    \setlength{\tabcolsep}{2.6pt}
    \caption{
        Validation-time evidence harnesses.
        \textit{Seg.} denotes segmentation outlines with displayed cell identifiers.
        Zoomed crops are either provided directly or obtained through the indicated tool.
        The two rightmost columns give NN-residual F1 (mean $\pm$ standard error over the
        178 items with residual links) on the hard validation split (single completion,
        frozen budget); \texttt{claude-opus-4.8} was run post hoc on c1/c4/c8 only.
    }
    \label{tab:evidence_harnesses}

    \resizebox{\textwidth}{!}{%
    \begin{tabular}{c cccc cccc c l l rr}
        \toprule
        & \multicolumn{4}{c}{Frame $t$}
        & \multicolumn{4}{c}{Frame $t+1$}
        & & &
        & \multicolumn{2}{c}{NN-res.\ F1} \\
        \cmidrule(lr){2-5}
        \cmidrule(lr){6-9}
        \cmidrule(lr){13-14}

        Condition
        & Raw
        & BBox
        & Seg.
        & Zoomed crops
        & Raw
        & BBox
        & Seg.
        & Zoomed crops
        & Multiturn
        & Tool
        & Purpose
        & \texttt{gpt-5.5}
        & \texttt{opus-4.8} \\

        \midrule

        c1
        & \checkmark & \checkmark & -- & --
        & \checkmark & \checkmark & -- & --
        & \checkmark
        & \texttt{crop\_and\_zoom(...)}
        & Coordinate baseline
        & $0.497{\scriptstyle\pm.033}$ & $0.216{\scriptstyle\pm.028}$ \\

        c2
        & \checkmark & -- & \checkmark & --
        & \checkmark & -- & \checkmark & --
        & \checkmark
        & \texttt{crop\_and\_zoom(...)}
        & Visual localization
        & $0.329{\scriptstyle\pm.027}$ & --- \\

        c3
        & \checkmark & -- & \checkmark & --
        & \checkmark & -- & \checkmark & --
        & \checkmark
        & \texttt{cell\_region(cell\_id: int)}
        & Cell-addressed inspection
        & $0.347{\scriptstyle\pm.030}$ & --- \\

        c4
        & \checkmark & -- & \checkmark & Outlined
        & \checkmark & -- & \checkmark & Outlined
        & --
        & --
        & Fixed visual inspection
        & $0.351{\scriptstyle\pm.029}$ & $0.150{\scriptstyle\pm.022}$ \\

        c5
        & \checkmark & -- & \checkmark & Plain
        & \checkmark & -- & \checkmark & Plain
        & --
        & --
        & Crop-outline ablation
        & $0.331{\scriptstyle\pm.028}$ & --- \\

        c6
        & \checkmark & \checkmark & -- & --
        & \checkmark & -- & \checkmark & --
        & \checkmark
        & \texttt{crop\_and\_zoom(...)}
        & Target-coordinate ablation
        & $0.342{\scriptstyle\pm.028}$ & --- \\

        c7
        & -- & -- & -- & Outlined
        & -- & -- & -- & Outlined
        & --
        & --
        & Global-position ablation
        & $0.151{\scriptstyle\pm.021}$ & --- \\

        c8
        & \checkmark & \checkmark & \checkmark & --
        & \checkmark & \checkmark & \checkmark & --
        & \checkmark
        & \texttt{crop\_and\_zoom(...)}
        & Combined evidence
        & $0.479{\scriptstyle\pm.033}$ & $0.248{\scriptstyle\pm.027}$ \\

        \bottomrule
    \end{tabular}%
    }
\end{table*}

The harness selection uses only the validation partition that is disjoint in 
sequence. We assess all eight conditions on the 205 validation neighborhoods 
and all three tasks using \texttt{gpt-5.5} with a low level of reasoning 
effort, which yields 4,920 turn-based conversations considering all problems, 
tasks, and conditions. Each request is given a total budget of 
8,192 tokens for both reasoning and tool use, 2,048 of which are set aside for 
the final structured answer. The selection statistic, reported in Appendix 
Table~\ref{tab:harness_validation}, is the average of the three dataset-macro 
Overall F1 scores;
not to be confused with the diagnostic NN-residual linking F1 shown in 
Table~\ref{tab:evidence_harnesses}.
By this selection statistic, condition c1 
has the highest validation mean, 0.609, compared with 0.547 for the next-best 
condition, c8. The difference is 0.061, and the 95\% 
hierarchical-bootstrap confidence interval is [0.005,0.120]. We therefore 
chose c1, where the model is given an unannotated overview of the two frames 
and the 
normalized bounding boxes for each candidate cell, and it can obtain closer 
views by using \texttt{crop\_and\_zoom}. A post-hoc check carried out with 
\texttt{claude-opus-4.8} confirms the advantage of tool-bearing coordinate 
interfaces over the evidence-supplied interface, even though it does not 
independently identify c1 as being better than c8.

We freeze the renderer, prompt, tool implementation, parser, token budget, and 
decoding configuration after selection. Each model included on the leaderboard 
is evaluated using the same c1 harness without making any further interface 
choices, and the test results take no part in 
the selection process. Detailed versions of the prompt and tool schema are 
given in Appendices~\ref{app:c1_prompt} and~\ref{app:tool_schema}.

\section{Results and Discussion}
\label{sec:results}

\begin{table*}[t]
    \centering
    \scriptsize
    \setlength{\tabcolsep}{2.3pt}
    \renewcommand{\arraystretch}{0.94}
    \caption{Unified frozen-c1 hard-test results. Primary task scores are shaded blue,
    proximity-resistance metrics green, and graph-level repair and inspection utility peach.
    LNK$_{\mathrm{sub}}$ is the graph-edit score, $\Delta$NN is its margin over the
    nearest-centroid graph, and $\Delta$/1k is inspection gain per 1{,}000 flagged edges.
    Reference graphs and trackers are evaluated only where their outputs define the metric.}
    \label{tab:results_merged_wide}
    \resizebox{\textwidth}{!}{%
    \begin{tabular}{
        lc
        >{\columncolor{primarypastel}}r
        >{\columncolor{primarypastel}}r
        >{\columncolor{primarypastel}}r
        >{\columncolor{primarypastel}}r
        >{\columncolor{resistancepastel}}r
        >{\columncolor{resistancepastel}}r
        >{\columncolor{resistancepastel}}r
        >{\columncolor{resistancepastel}}r
        >{\columncolor{utilitypastel}}r
        >{\columncolor{utilitypastel}}r
        >{\columncolor{utilitypastel}}r
    }
        \toprule
        System & Access
        & \multicolumn{4}{c}{\cellcolor{primarypastel}\textbf{Primary task scores}}
        & \multicolumn{4}{c}{\cellcolor{resistancepastel}\textbf{Shortcut resistance}}
        & \multicolumn{3}{c}{\cellcolor{utilitypastel}\textbf{Graph utility}} \\
        \cmidrule(lr){3-6}\cmidrule(lr){7-10}\cmidrule(lr){11-13}
        & & Link F1 & Repair F1 & Inspect F1 & Mean
        & NN-res & Div & LNK$_{\mathrm{sub}}$ & $\Delta$NN
        & Repair $\Delta$LNK & Inspect $\Delta$LNK & $\Delta$/1k \\
        \midrule
        GPT-5.6 (sol)~\cite{openai2026gpt56} & API & 0.811 & 0.815 & 0.631 & 0.752 & 0.723 & 0.787 & 0.767 & $+0.274$ & $+0.412$ & $+0.518$ & 0.555 \\
        GPT-5.6 (terra)~\cite{openai2026gpt56} & API & 0.792 & 0.788 & 0.590 & 0.724 & 0.634 & 0.707 & 0.692 & $+0.199$ & $+0.337$ & $+0.518$ & 0.529 \\
        Claude Fable 5~\cite{anthropic2026fable5} & API & 0.790 & 0.786 & 0.698 & 0.758 & 0.521 & 0.573 & 0.664 & $+0.171$ & $+0.348$ & $+0.513$ & 0.517 \\
        GPT-5.5~\cite{openai2026gpt55} & API & 0.717 & 0.741 & 0.588 & 0.682 & 0.577 & 0.622 & 0.638 & $+0.145$ & $+0.280$ & $+0.500$ & 0.511 \\
        Claude Opus 5~\cite{anthropic2026opus5} & API & 0.758 & 0.739 & 0.532 & 0.676 & 0.406 & 0.411 & 0.623 & $+0.130$ & $+0.263$ & $+0.459$ & 0.542 \\
        Claude Opus 4.8~\cite{anthropic2026opus48} & API & 0.653 & 0.729 & 0.499 & 0.627 & 0.186 & 0.181 & 0.502 & $+0.009$ & $+0.201$ & $+0.437$ & 0.504 \\
        \midrule
        \textbf{Nearest-centroid graph} & --- & --- & --- & --- & --- & --- & --- & 0.493 & --- & --- & --- & --- \\
        \textbf{Hungarian matcher} & --- & 0.681 & --- & --- & --- & --- & --- & 0.492 & --- & --- & --- & --- \\
        \textbf{Trackastra}~\cite{gallusser2024trackastra} & --- & 0.698 & --- & --- & --- & --- & --- & 0.484 & --- & --- & --- & --- \\
        \midrule
        Gemma 4 31B~\cite{gemmateam2026gemma4} & Weights & 0.641 & 0.606 & 0.267 & 0.505 & 0.149 & 0.103 & 0.484 & $-0.009$ & $+0.030$ & $+0.314$ & 0.520 \\
        GPT-5.6 (luna)~\cite{openai2026gpt56} & API & 0.655 & 0.697 & 0.359 & 0.570 & 0.517 & 0.534 & 0.481 & $-0.012$ & $+0.180$ & $+0.463$ & 0.504 \\
        Claude Haiku 4.5~\cite{anthropic2025haiku45} & API & 0.613 & 0.629 & 0.266 & 0.503 & 0.169 & 0.137 & 0.450 & $-0.042$ & $+0.097$ & $+0.307$ & 0.565 \\
        Qwen3.5-397B-A17B~\cite{qwen2026qwen35} & Weights & 0.419 & 0.656 & 0.377 & 0.484 & 0.241 & 0.274 & 0.444 & $-0.049$ & $+0.166$ & $+0.342$ & 0.545 \\
        Qwen3.5-122B-A10B~\cite{qwen2026qwen35} & Weights & 0.503 & 0.598 & 0.227 & 0.443 & 0.118 & 0.104 & 0.437 & $-0.055$ & $+0.055$ & $+0.322$ & 0.521 \\
        \midrule
        \textbf{HOCT}~\cite{bragantini2026higher} & --- & 0.622 & --- & --- & --- & --- & --- & 0.385 & --- & --- & --- & --- \\
        \midrule
        GLM-4.6V~\cite{glmteam2025glm45v} & Weights & 0.454 & 0.519 & 0.183 & 0.385 & 0.091 & 0.053 & 0.338 & $-0.155$ & $-0.094$ & $+0.318$ & 0.444 \\
        Inkling-Small~\cite{thinkingmachines2026inklingsmall} & Weights & 0.538 & 0.443 & 0.195 & 0.392 & 0.181 & 0.143 & 0.314 & $-0.179$ & $-0.332$ & $+0.521$ & 0.378 \\
        MiniMax M3~\cite{minimax2026m3} & Weights & 0.161 & 0.656 & 0.183 & 0.333 & 0.237 & 0.290 & 0.300 & $-0.192$ & $+0.139$ & $+0.209$ & 0.562 \\
        Claude Sonnet 5~\cite{anthropic2026sonnet5} & API & 0.515 & 0.631 & 0.079 & 0.409 & 0.157 & 0.117 & 0.281 & $-0.211$ & $+0.010$ & $+0.174$ & 0.472 \\
        Qwen3.6-27B~\cite{qwen2026qwen36} & Weights & 0.229 & 0.631 & 0.119 & 0.326 & 0.093 & 0.094 & 0.197 & $-0.295$ & $+0.055$ & $+0.105$ & 0.468 \\
        Qwen3.5-35B-A3B~\cite{qwen2026qwen35} & Weights & 0.096 & 0.622 & 0.076 & 0.264 & 0.098 & 0.121 & 0.147 & $-0.345$ & $+0.008$ & $+0.094$ & 0.498 \\
        Kimi K3~\cite{kimiteam2026k3} & Weights & 0.087 & 0.659 & 0.083 & 0.276 & 0.117 & 0.135 & 0.144 & $-0.348$ & $+0.111$ & $+0.128$ & 0.556 \\
        InternVL3.5-38B~\cite{wang2025internvl35} & Weights & 0.156 & 0.556 & 0.026 & 0.246 & 0.031 & 0.025 & 0.091 & $-0.402$ & $-0.085$ & $+0.129$ & 0.419 \\
        Mistral Medium 3.5 128B~\cite{mistral2026medium35} & Weights & 0.262 & 0.465 & 0.084 & 0.270 & 0.116 & 0.097 & 0.085 & $-0.408$ & $-0.244$ & $+0.216$ & 0.216 \\
        Molmo2-8B~\cite{clark2026molmo2} & Weights & 0.077 & 0.521 & 0.015 & 0.204 & 0.018 & 0.004 & 0.000 & $-0.493$ & $-0.099$ & $+0.367$ & 0.511 \\
        Llama 4 Scout~\cite{meta2025llama4} & Weights & 0.387 & 0.342 & 0.098 & 0.275 & 0.091 & 0.044 & 0.000 & $-0.493$ & $-0.385$ & $+0.519$ & 0.170 \\
        \bottomrule
    \end{tabular}%
    }
\end{table*}

\textbf{Leaderboard and shortcut resistance.} The primary, shortcut-resistance, 
and graph-level results are summarized in Table~\ref{tab:results_merged_wide}. 
The nearest-centroid baseline achieves LNK$_{\mathrm{sub}}$ of 0.493, showing 
that cell linking can already be solved relatively well by proximity alone. The 
only models to overcome this baseline are GPT-5.6 (sol), with $0.767$
($\Delta$NN $+0.274$), followed by GPT-5.6 (terra), Claude Fable 5, GPT-5.5, 
and Claude Opus 5; and Claude Opus 4.8, which is nearly equal to the NN 
baseline (+0.009). All open-weight models stay below it.

The aggregate linking F1 can shadow real deficiencies, justifying the proposed 
metrics.
For example, Gemma 4 31B is the most powerful open-weight model with a Link F1 of
0.641, 
almost matching the specialized trackers, but its LNK$_{\mathrm{sub}}$ is 
0.484, and its NN-residual and division F1 scores are only 0.149 and 0.103, 
respectively.
This suggests that this model and other open-weight models perform well because 
they rely on proximity for matching. In contrast, the best-performing model on
these metrics, GPT-5.6 (sol),
scores an NN-residual F1 of 0.723 and a division F1 of 0.787. It is important 
to note that low tracker scores do not indicate anything about their 
full-sequence performance, since the benchmark problems were deliberately 
chosen to be ones where these trackers fail.

\textbf{Can VLMs repair tracker output?}
Appendix Table~\ref{tab:repair_breakdown} decomposes repair behavior by edit 
type and per-neighborhood graph. Repair F1 measures how much of the proposed
graph is correct after editing, but not whether the model improves the graph.
For that, we rely on $\Delta$LNK. GPT-5.6 (sol) is the most useful, raising LNK 
by +0.412 and 
improving 80\% of the cases while worsening only 7\% and 13\% kept unchanged.
On the other hand, Qwen3.5-35B-A3B achieves a repair F1 of 0.622, but its LNK is
increased only by +0.008, 
mostly because it doesn't change the proposed graph. Even more, Molmo2-8B and
Llama 4 Scout have repair F1 scores of 0.521 and 0.342, respectively, while
changing LNK by -0.099 and -0.385. Six of the 13 open-weight models actually
hurt performance and give a negative $\Delta$LNK. For that reason, repair must be
interpreted by graph improvement, edit precision, and recall.

\textbf{Can VLMs prioritize human review?}
Appendix Table~\ref{tab:triage_budget} evaluates inspection under fixed review budgets.
Inspection is only useful if it identifies errors early enough to reduce human effort.
We therefore cap flags at a fixed review budget and simulate the correction of the selected sources.
Under the most restrictive budget, the frontier models 
achieve a recovery of $\Delta$LNK between +0.235 and +0.249, which is 83--88\% 
of the gold oracle's +0.284 and almost twice that of a random reviewer 
(+0.125). However, the model can produce indiscriminate flags. For example, 
Llama 4 Scout produces 3,061 flags and recovers +0.519, but it achieves only
+0.117 when the tighter budget is applied, which is below random selection. Therefore, inspection must be analyzed according to the review budget, and not only by the improvement when examining all the flags.

\textbf{Difficulty factors and task relationships.} Proximity resistance is the 
hardest challenge. The nearest-centroid rule solves just 22 out of 244 problems 
exactly. All models do well on these 22, but their performance drops on the 
other 222. For these NN-resistant cases, LNK$_{\mathrm{sub}}$ is $0.766$ for 
GPT-5.6 (sol) and $0.536$ for Gemma 4 31B. Weaker models also struggle more with
divisions.

The mining-hardness score shows another side of difficulty
(Appendix Table~\ref{tab:difficulty_strata}).
Performance usually drops as we move to harder groups, but not for every model. 
The hard-mining procedure shapes how the benchmark is chosen: all three trackers
agree on 235 out of 244 test neighborhoods. This means the benchmark focuses on 
consistently tough cases. Because of this, there is little variation, so 
proximity resistance and division metrics help sort the cases more clearly.

The three tasks are related but have different roles. Linking and repair scores 
are 
closely tied across models (Spearman $\rho=0.85$), so models that rebuild 
associations well also fix them well. Inspection is less connected, with a 
Spearman of $\rho=0.51$ with linking and $\rho=0.29$ with repair. This shows 
that finding links to review is a separate skill. Each task, therefore, 
highlights a unique way a VLM can support the tracking process.

\textbf{Interface and inference efficiency.} The harness selection study 
benefited from coordinate-based interfaces with model-requested crops in a 
multi-turn fashion over fixed supplied crops in a single-turn fashion.  For 
GPT-5.5, c1 performs the best; meanwhile, Claude Opus 4.8 favors the more 
information-rich c8 configuration. This model-dependent order suggests that 
both harnesses provide access to the visual evidence effectively; meanwhile, 
different models use the additional segmentation delineation cues differently. 
We selected c1 and froze it for the final testing, ensuring that every model is 
evaluated equally.
A validation-test comparison under the frozen protocol is reported in Appendix 
Table~\ref{tab:validation_robustness}.

Moreover, through a fixed inference budget, we measured how effectively a model 
can complete the task within a reasonable amount of time.
This provides interesting insights into some open-weight models.
For example, Qwen3.5-35B-A3B responses are truncated in 87\% of problems, even
with the final low-context warning. In contrast, larger variants are truncated 
by $\approx$40\%. This reflects inference efficiency, a key concern in a 
tool-using setting. %

\section{Conclusion}
\label{sec:conclusion}

We introduced CellTrack-Bench, a benchmark for assessing vision-language models 
(VLMs) for cell tracking by breaking long tracking sequences into compact,
digestible tasks for current models. Given the skewed distribution of mistakes, 
we selected difficult associations from common mistakes made by specialized 
trackers and formulated three complementary tasks: cell linking, track repair, 
and track inspection. Our formulation lets us evaluate whether VLMs can 
directly restore cell associations, fix existing tracking solutions, or
prioritize likely mistakes for review during human data curation.

Our results suggest that advanced VLMs possess abilities that go beyond mere 
geometric tricks. The most advanced VLMs, the closed-source frontier models, 
can operate without nearest-neighbor shortcuts, improve poorly performing 
tracker solutions, and prioritize tracker errors even under limited inspection
budgets. However, these abilities are not universal: weaker models
still rely on the proximity shortcut and, in some cases, can even worsen 
tracking performance during repair. Taken together, these facts suggest that, 
for now, VLMs are better used as reasoning modules in the cell-tracking 
pipeline rather than as substitutes for specialized trackers. We expect our 
CellTrack-Bench to serve as an evaluation setting for developing VLMs capable of cognitive reasoning about complex biological events, where  specialized trackers often make mistakes.

\section*{LLM Disclosure}
We used large language model assistants, including Claude and GPT-series 
models, to help draft and revise prose, develop and debug the evaluation and 
analysis code, and create figure-generation scripts. The authors designed the 
study, chose the experimental and statistical procedures, ran the experiments, 
and reviewed and verified all model-assisted text, code, analyses, figures, and 
reported results. The authors take full responsibility for the paper's content.

\section*{Licensing Disclosure}

The CTC and bacterial datasets remain subject to their own licenses, which do not allow us
to redistribute them or annotations derived from them, so we do not distribute the benchmark
items. Instead, we release the code, the stored model outputs, and the saved tracker outputs
together with a construction pipeline that regenerates the benchmark from the original
datasets and verifies it, file by file, against the version evaluated here.

{
    \small
    \bibliographystyle{ieeenat_fullname}
    \bibliography{main}

\begin{thebibliography}{48}
\providecommand{\natexlab}[1]{#1}
\providecommand{\url}[1]{\texttt{#1}}
\expandafter\ifx\csname urlstyle\endcsname\relax
  \providecommand{\doi}[1]{doi: #1}\else
  \providecommand{\doi}{doi: \begingroup \urlstyle{rm}\Url}\fi

\bibitem[{Anthropic}(2025)]{anthropic2025haiku45}
{Anthropic}.
\newblock Introducing claude haiku 4.5.
\newblock \url{https://www.anthropic.com/news/claude-haiku-4-5}, 2025.
\newblock Accessed: 2026-08-28.

\bibitem[{Anthropic}(2026{\natexlab{a}})]{anthropic2026fable5}
{Anthropic}.
\newblock Claude fable 5 and claude mythos 5.
\newblock \url{https://www.anthropic.com/news/claude-fable-5-mythos-5},
  2026{\natexlab{a}}.
\newblock Accessed: 2026-08-28.

\bibitem[{Anthropic}(2026{\natexlab{b}})]{anthropic2026opus48}
{Anthropic}.
\newblock Introducing claude opus 4.8.
\newblock \url{https://www.anthropic.com/news/claude-opus-4-8},
  2026{\natexlab{b}}.
\newblock Accessed: 2026-08-28.

\bibitem[{Anthropic}(2026{\natexlab{c}})]{anthropic2026opus5}
{Anthropic}.
\newblock Introducing claude opus 5.
\newblock \url{https://www.anthropic.com/news/claude-opus-5},
  2026{\natexlab{c}}.
\newblock Accessed: 2026-08-28.

\bibitem[{Anthropic}(2026{\natexlab{d}})]{anthropic2026sonnet5}
{Anthropic}.
\newblock Introducing claude sonnet 5.
\newblock \url{https://www.anthropic.com/news/claude-sonnet-5},
  2026{\natexlab{d}}.
\newblock Accessed: 2026-08-28.

\bibitem[Betjes et~al.(2025)Betjes, Kok, Tans, and van Zon]{betjes2025cell}
Max~A Betjes, Rutger~NU Kok, Sander~J Tans, and Jeroen~S van Zon.
\newblock Cell tracking with accurate error prediction.
\newblock \emph{Nature Methods}, 22\penalty0 (11):\penalty0 2400--2410, 2025.

\bibitem[Bragantini et~al.(2025)Bragantini, Theodoro, Zhao, Huijben,
  Hirata-Miyasaki, VijayKumar, Balasubramanian, Lao, Agrawal, Xiao,
  et~al.]{bragantini2025ultrack}
Jord{\~a}o Bragantini, Ilan Theodoro, Xiang Zhao, Teun~APM Huijben, Eduardo
  Hirata-Miyasaki, Shruthi VijayKumar, Akilandeswari Balasubramanian, Tiger
  Lao, Richa Agrawal, Sheng Xiao, et~al.
\newblock Ultrack: pushing the limits of cell tracking across biological
  scales.
\newblock \emph{Nature Methods}, 22\penalty0 (11):\penalty0 2423--2436, 2025.

\bibitem[Bragantini et~al.(2026)Bragantini, Theodoro, and
  Royer]{bragantini2026higher}
Jord{\~a}o Bragantini, Ilan Theodoro, and Lo{\"\i}c~A Royer.
\newblock Higher-order cell tracking transformer.
\newblock \emph{arXiv preprint arXiv:2607.11754}, 2026.

\bibitem[Burgess et~al.(2025)Burgess, Nirschl, Bravo-S{\'a}nchez, Lozano,
  Gupte, Galaz-Montoya, Zhang, Su, Bhowmik, Coman, et~al.]{burgess2025microvqa}
James Burgess, Jeffrey~J Nirschl, Laura Bravo-S{\'a}nchez, Alejandro Lozano,
  Sanket~Rajan Gupte, Jesus~G Galaz-Montoya, Yuhui Zhang, Yuchang Su, Disha
  Bhowmik, Zachary Coman, et~al.
\newblock Microvqa: A multimodal reasoning benchmark for microscopy-based
  scientific research.
\newblock In \emph{2025 IEEE/CVF Conference on Computer Vision and Pattern
  Recognition (CVPR)}, pages 19553--19564. IEEE, 2025.

\bibitem[Chen et~al.(2024)Chen, Li, Dong, Zhang, Zang, Chen, Duan, Wang, Qiao,
  Lin, et~al.]{chen2024we}
Lin Chen, Jinsong Li, Xiaoyi Dong, Pan Zhang, Yuhang Zang, Zehui Chen, Haodong
  Duan, Jiaqi Wang, Yu Qiao, Dahua Lin, et~al.
\newblock Are we on the right way for evaluating large vision-language models?
\newblock \emph{Advances in Neural Information Processing Systems},
  37:\penalty0 27056--27087, 2024.

\bibitem[Clark et~al.(2026)Clark, Zhang, Ma, Park, Salehi, Tripathi, Lee, Ren,
  Kim, Yang, Shao, Yang, Huang, Gao, Anderson, Zhang, Jain, Stoica, Han,
  Farhadi, and Krishna]{clark2026molmo2}
Christopher Clark, Jieyu Zhang, Zixian Ma, Jae~Sung Park, Mohammadreza Salehi,
  Rohun Tripathi, Sangho Lee, Zhongzheng Ren, Chris~Dongjoo Kim, Yinuo Yang,
  Vincent Shao, Yue Yang, Weikai Huang, Ziqi Gao, Taira Anderson, Jianrui
  Zhang, Jitesh Jain, George Stoica, Winson Han, Ali Farhadi, and Ranjay
  Krishna.
\newblock {Molmo2}: Open weights and data for vision-language models with video
  understanding and grounding.
\newblock \emph{arXiv preprint arXiv:2601.10611}, 2026.

\bibitem[Dendorfer et~al.(2020)Dendorfer, Rezatofighi, Milan, Shi, Cremers,
  Reid, Roth, Schindler, and Leal-Taix{\'e}]{dendorfer2020mot20}
Patrick Dendorfer, Hamid Rezatofighi, Anton Milan, Javen Shi, Daniel Cremers,
  Ian Reid, Stefan Roth, Konrad Schindler, and Laura Leal-Taix{\'e}.
\newblock Mot20: A benchmark for multi object tracking in crowded scenes.
\newblock \emph{arXiv preprint arXiv:2003.09003}, 2020.

\bibitem[Fukai and Kawaguchi(2023)]{fukai2023laptrack}
Yohsuke~T Fukai and Kyogo Kawaguchi.
\newblock Laptrack: linear assignment particle tracking with tunable metrics.
\newblock \emph{Bioinformatics}, 39\penalty0 (1):\penalty0 btac799, 2023.

\bibitem[Gallusser and Weigert(2024)]{gallusser2024trackastra}
Benjamin Gallusser and Martin Weigert.
\newblock Trackastra: Transformer-based cell tracking for live-cell microscopy.
\newblock In \emph{European conference on computer vision}, pages 467--484.
  Springer, 2024.

\bibitem[Geirhos et~al.(2020)Geirhos, Jacobsen, Michaelis, Zemel, Brendel,
  Bethge, and Wichmann]{geirhos2020shortcut}
Robert Geirhos, J{\"o}rn-Henrik Jacobsen, Claudio Michaelis, Richard Zemel,
  Wieland Brendel, Matthias Bethge, and Felix~A Wichmann.
\newblock Shortcut learning in deep neural networks.
\newblock \emph{Nature Machine Intelligence}, 2\penalty0 (11):\penalty0
  665--673, 2020.

\bibitem[{Gemma Team} et~al.(2026)]{gemmateam2026gemma4}
{Gemma Team} et~al.
\newblock Gemma 4 technical report.
\newblock \emph{arXiv preprint arXiv:2607.02770}, 2026.

\bibitem[{GLM-V Team}(2025)]{glmteam2025glm45v}
{GLM-V Team}.
\newblock {GLM-4.5V} and {GLM-4.1V-Thinking}: Towards versatile multimodal
  reasoning with scalable reinforcement learning.
\newblock \emph{arXiv preprint arXiv:2507.01006}, 2025.

\bibitem[Goldsborough et~al.(2024)Goldsborough, Philps, O'Callaghan, Inglis,
  Leplat, Filby, Bilen, and Bankhead]{goldsborough2024instanseg}
Thibaut Goldsborough, Ben Philps, Alan O'Callaghan, Fiona Inglis, Leo Leplat,
  Andrew Filby, Hakan Bilen, and Peter Bankhead.
\newblock Instanseg: an embedding-based instance segmentation algorithm
  optimized for accurate, efficient and portable cell segmentation.
\newblock \emph{arXiv preprint arXiv:2408.15954}, 2024.

\bibitem[Hu et~al.(2024)Hu, Li, Lu, Shao, He, Qiao, and Luo]{hu2024omnimedvqa}
Yutao Hu, Tianbin Li, Quanfeng Lu, Wenqi Shao, Junjun He, Yu Qiao, and Ping
  Luo.
\newblock Omnimedvqa: A new large-scale comprehensive evaluation benchmark for
  medical lvlm.
\newblock In \emph{2024 IEEE/CVF Conference on Computer Vision and Pattern
  Recognition (CVPR)}, pages 22170--22183. IEEE, 2024.

\bibitem[Ichbiah et~al.(2023)Ichbiah, Delbary, McDougall, Dumollard, and
  Turlier]{ichbiah2023embryo}
Sacha Ichbiah, Fabrice Delbary, Alex McDougall, R{\'e}mi Dumollard, and
  Herv{\'e} Turlier.
\newblock Embryo mechanics cartography: inference of 3d force atlases from
  fluorescence microscopy.
\newblock \emph{Nature methods}, 20\penalty0 (12):\penalty0 1989--1999, 2023.

\bibitem[Jaqaman et~al.(2008)Jaqaman, Loerke, Mettlen, Kuwata, Grinstein,
  Schmid, and Danuser]{jaqaman2008robust}
Khuloud Jaqaman, Dinah Loerke, Marcel Mettlen, Hirotaka Kuwata, Sergio
  Grinstein, Sandra~L Schmid, and Gaudenz Danuser.
\newblock Robust single-particle tracking in live-cell time-lapse sequences.
\newblock \emph{Nature methods}, 5\penalty0 (8):\penalty0 695--702, 2008.

\bibitem[{Kimi Team}(2026)]{kimiteam2026k3}
{Kimi Team}.
\newblock Kimi k3: Open frontier intelligence.
\newblock \emph{arXiv preprint arXiv:2607.24653}, 2026.

\bibitem[Kok et~al.(2020)Kok, Hebert, Huelsz-Prince, Goos, Zheng, Bozek,
  Stephens, Tans, and Van~Zon]{kok2020organoidtracker}
Rutger~NU Kok, Laetitia Hebert, Guizela Huelsz-Prince, Yvonne~J Goos, Xuan
  Zheng, Katarzyna Bozek, Greg~J Stephens, Sander~J Tans, and Jeroen~S Van~Zon.
\newblock Organoidtracker: Efficient cell tracking using machine learning and
  manual error correction.
\newblock \emph{PLoS One}, 15\penalty0 (10):\penalty0 e0240802, 2020.

\bibitem[Lange et~al.(2024)Lange, Granados, VijayKumar, Bragantini, Ancheta,
  Kim, Santhosh, Borja, Kobayashi, McGeever, et~al.]{lange2024multimodal}
Merlin Lange, Alejandro Granados, Shruthi VijayKumar, Jord{\~a}o Bragantini,
  Sarah Ancheta, Yang-Joon Kim, Sreejith Santhosh, Michael Borja, Hirofumi
  Kobayashi, Erin McGeever, et~al.
\newblock A multimodal zebrafish developmental atlas reveals the
  state-transition dynamics of late-vertebrate pluripotent axial progenitors.
\newblock \emph{Cell}, 187\penalty0 (23):\penalty0 6742--6759, 2024.

\bibitem[Laviron et~al.(2019)Laviron, Combadi{\`e}re, and
  Boissonnas]{laviron2019tracking}
Marie Laviron, Christophe Combadi{\`e}re, and Alexandre Boissonnas.
\newblock Tracking monocytes and macrophages in tumors with live imaging.
\newblock \emph{Frontiers in immunology}, 10:\penalty0 1201, 2019.

\bibitem[Li et~al.(2024)Li, Wang, He, Li, Wang, Liu, Wang, Xu, Chen, Luo,
  et~al.]{li2024mvbench}
Kunchang Li, Yali Wang, Yinan He, Yizhuo Li, Yi Wang, Yi Liu, Zun Wang, Jilan
  Xu, Guo Chen, Ping Luo, et~al.
\newblock Mvbench: A comprehensive multi-modal video understanding benchmark.
\newblock In \emph{Proceedings of the IEEE/CVF conference on computer vision
  and pattern recognition}, pages 22195--22206, 2024.

\bibitem[Lozano et~al.(2024)Lozano, Nirschl, Burgess, Gupte, Zhang, Unell, and
  Yeung-Levy]{lozano2024micro}
Alejandro Lozano, Jeffrey Nirschl, James Burgess, Sanket~R Gupte, Yuhui Zhang,
  Alyssa Unell, and Serena Yeung-Levy.
\newblock Micro-bench: A microscopy benchmark for vision-language
  understanding.
\newblock \emph{Advances in Neural Information Processing Systems},
  37:\penalty0 30670--30685, 2024.

\bibitem[Ma{\v{s}}ka et~al.(2023)Ma{\v{s}}ka, Ulman, Delgado-Rodriguez,
  G{\'o}mez-de Mariscal, Ne{\v{c}}asov{\'a}, Guerrero~Pe{\~n}a, Ren,
  Meyerowitz, Scherr, L{\"o}ffler, et~al.]{mavska2023cell}
Martin Ma{\v{s}}ka, Vladim{\'\i}r Ulman, Pablo Delgado-Rodriguez, Estibaliz
  G{\'o}mez-de Mariscal, Tereza Ne{\v{c}}asov{\'a}, Fidel~A Guerrero~Pe{\~n}a,
  Tsang~Ing Ren, Elliot~M Meyerowitz, Tim Scherr, Katharina L{\"o}ffler, et~al.
\newblock The cell tracking challenge: 10 years of objective benchmarking.
\newblock \emph{Nature methods}, 20\penalty0 (7):\penalty0 1010--1020, 2023.

\bibitem[Matula et~al.(2015)Matula, Ma{\v{s}}ka, Sorokin, Matula, Ortiz-de
  Sol{\'o}rzano, and Kozubek]{matula2015aogm}
Pavel Matula, Martin Ma{\v{s}}ka, Dmitry~V Sorokin, Petr Matula, Carlos
  Ortiz-de Sol{\'o}rzano, and Michal Kozubek.
\newblock Cell tracking accuracy measurement based on comparison of acyclic
  oriented graphs.
\newblock \emph{PloS one}, 10\penalty0 (12):\penalty0 e0144959, 2015.

\bibitem[McDole et~al.(2018)McDole, Guignard, Amat, Berger, Malandain, Royer,
  Turaga, Branson, and Keller]{mcdole2018toto}
Katie McDole, L{\'e}o Guignard, Fernando Amat, Andrew Berger, Gr{\'e}goire
  Malandain, Lo{\"\i}c~A Royer, Srinivas~C Turaga, Kristin Branson, and
  Philipp~J Keller.
\newblock In toto imaging and reconstruction of post-implantation mouse
  development at the single-cell level.
\newblock \emph{Cell}, 175\penalty0 (3):\penalty0 859--876, 2018.

\bibitem[Meijering et~al.(2009)Meijering, Dzyubachyk, Smal, and van
  Cappellen]{meijering2009tracking}
Erik Meijering, Oleh Dzyubachyk, Ihor Smal, and Wiggert~A van Cappellen.
\newblock Tracking in cell and developmental biology.
\newblock In \emph{Seminars in cell \& developmental biology}, pages 894--902.
  Elsevier, 2009.

\bibitem[{Meta AI}(2025)]{meta2025llama4}
{Meta AI}.
\newblock The llama 4 herd: The beginning of a new era of natively multimodal
  {AI} innovation.
\newblock \url{https://ai.meta.com/blog/llama-4-multimodal-intelligence/},
  2025.
\newblock Accessed: 2026-08-28.

\bibitem[{MiniMax}(2026)]{minimax2026m3}
{MiniMax}.
\newblock {MiniMax M3}: Frontier coding, 1m context, native multimodality---all
  in one model.
\newblock \url{https://www.minimax.io/blog/minimax-m3}, 2026.
\newblock Accessed: 2026-08-28.

\bibitem[{Mistral AI}(2026)]{mistral2026medium35}
{Mistral AI}.
\newblock Mistral medium 3.5.
\newblock \url{https://docs.mistral.ai/models/mistral-medium-3-5-26-04}, 2026.
\newblock Accessed: 2026-08-28.

\bibitem[{OpenAI}(2026{\natexlab{a}})]{openai2026gpt55}
{OpenAI}.
\newblock Introducing {GPT-5.5}.
\newblock \url{https://openai.com/index/introducing-gpt-5-5/},
  2026{\natexlab{a}}.
\newblock Accessed: 2026-08-28.

\bibitem[{OpenAI}(2026{\natexlab{b}})]{openai2026gpt56}
{OpenAI}.
\newblock {GPT-5.6}: Frontier intelligence that scales with your ambition.
\newblock \url{https://openai.com/index/gpt-5-6/}, 2026{\natexlab{b}}.
\newblock Accessed: 2026-08-28.

\bibitem[Pop et~al.(2025)Pop, Le~Bodic, and Nunez-Iglesias]{pop2025needles}
Draga~Doncila Pop, Pierre Le~Bodic, and Juan Nunez-Iglesias.
\newblock Needles in the haystack-rapid error detection for cell tracking
  solutions.
\newblock In \emph{2025 IEEE 22nd International Symposium on Biomedical Imaging
  (ISBI)}, pages 1--4. IEEE, 2025.

\bibitem[{Qwen Team}(2026{\natexlab{a}})]{qwen2026qwen35}
{Qwen Team}.
\newblock {Qwen3.5}: Towards native multimodal agents.
\newblock \url{https://qwen.ai/blog?id=qwen3.5}, 2026{\natexlab{a}}.
\newblock Accessed: 2026-08-28.

\bibitem[{Qwen Team}(2026{\natexlab{b}})]{qwen2026qwen36}
{Qwen Team}.
\newblock {Qwen3.6-27B}: Flagship-level coding in a 27b dense model.
\newblock \url{https://qwen.ai/blog?id=qwen3.6-27b}, 2026{\natexlab{b}}.
\newblock Accessed: 2026-08-28.

\bibitem[Schmidt et~al.(2018)Schmidt, Weigert, Broaddus, and
  Myers]{schmidt2018cell}
Uwe Schmidt, Martin Weigert, Coleman Broaddus, and Gene Myers.
\newblock Cell detection with star-convex polygons.
\newblock In \emph{International conference on medical image computing and
  computer-assisted intervention}, pages 265--273. Springer, 2018.

\bibitem[Stringer et~al.(2021)Stringer, Wang, Michaelos, and
  Pachitariu]{stringer2021cellpose}
Carsen Stringer, Tim Wang, Michalis Michaelos, and Marius Pachitariu.
\newblock Cellpose: a generalist algorithm for cellular segmentation.
\newblock \emph{Nature methods}, 18\penalty0 (1):\penalty0 100--106, 2021.

\bibitem[{Thinking Machines Lab}(2026)]{thinkingmachines2026inklingsmall}
{Thinking Machines Lab}.
\newblock Inkling-small model card.
\newblock \url{https://thinkingmachines.ai/model-card/inkling-small/}, 2026.
\newblock Accessed: 2026-08-28.

\bibitem[Tinevez et~al.(2017)Tinevez, Perry, Schindelin, Hoopes, Reynolds,
  Laplantine, Bednarek, Shorte, and Eliceiri]{tinevez2017trackmate}
Jean-Yves Tinevez, Nick Perry, Johannes Schindelin, Genevieve~M Hoopes,
  Gregory~D Reynolds, Emmanuel Laplantine, Sebastian~Y Bednarek, Spencer~L
  Shorte, and Kevin~W Eliceiri.
\newblock Trackmate: An open and extensible platform for single-particle
  tracking.
\newblock \emph{Methods}, 115:\penalty0 80--90, 2017.

\bibitem[Ulicna et~al.(2021)Ulicna, Vallardi, Charras, and
  Lowe]{ulicna2021automated}
Kristina Ulicna, Giulia Vallardi, Guillaume Charras, and Alan~R Lowe.
\newblock Automated deep lineage tree analysis using a bayesian single cell
  tracking approach.
\newblock \emph{Frontiers in Computer Science}, 3:\penalty0 734559, 2021.

\bibitem[Ulman et~al.(2017)Ulman, Ma{\v{s}}ka, Magnusson, Ronneberger, Haubold,
  Harder, Matula, Matula, Svoboda, Radojevic, et~al.]{ulman2017objective}
Vladim{\'\i}r Ulman, Martin Ma{\v{s}}ka, Klas~EG Magnusson, Olaf Ronneberger,
  Carsten Haubold, Nathalie Harder, Pavel Matula, Petr Matula, David Svoboda,
  Miroslav Radojevic, et~al.
\newblock An objective comparison of cell-tracking algorithms.
\newblock \emph{Nature methods}, 14\penalty0 (12):\penalty0 1141--1152, 2017.

\bibitem[Wang et~al.(2025)Wang, Gao, Gu, Pu, Cui, Wei, Liu, Jing, Ye, Shao,
  et~al.]{wang2025internvl35}
Weiyun Wang, Zhangwei Gao, Lixin Gu, Hengjun Pu, Long Cui, Xingguang Wei,
  Zhaoyang Liu, Linglin Jing, Shenglong Ye, Jie Shao, et~al.
\newblock {InternVL3.5}: Advancing open-source multimodal models in
  versatility, reasoning, and efficiency.
\newblock \emph{arXiv preprint arXiv:2508.18265}, 2025.

\bibitem[Yu et~al.(2025)Yu, Wang, Wu, Luo, Wang, Xie, Rajpurkar, Yang, Yang,
  Wang, et~al.]{yu2025medframeqa}
Suhao Yu, Haojin Wang, Juncheng Wu, Luyang Luo, Jingshen Wang, Cihang Xie,
  Pranav Rajpurkar, Carl Yang, Yang Yang, Kang Wang, et~al.
\newblock Medframeqa: A multi-image medical vqa benchmark for clinical
  reasoning.
\newblock \emph{arXiv preprint arXiv:2505.16964}, 2025.

\bibitem[Zeng et~al.(2026)Zeng, Wang, Ma, Yin, Wang, Liu, Kanazawa, Ni, Li,
  Zanette, and Feng]{schema2026}
Guanning Zeng, Jiani Wang, Wenjie Ma, Shaofeng Yin, Chenyang Wang, Shichen Liu,
  Angjoo Kanazawa, Wode Ni, Xiuyu Li, Andrea Zanette, and Haiwen Feng.
\newblock {[schema]: Frontier Models with the Right Harness Achieve $\sim$99\%
  on ARC-AGI-3 Public}.
\newblock Impossible Research. \url{https://schema-harness.github.io/}, 2026.

\end{thebibliography}
}

\clearpage
\appendix

\section{Benchmark Construction and Protocol Detail}
\label{app:benchmark-construction}
\label{app:data}

\subsection{Source Data and Sequence Splits}

CellTrack-Bench draws from 10 Cell Tracking Challenge (CTC) datasets and five
bacterial datasets, comprising 35 image sequences across fluorescence,
phase-contrast, differential-interference-contrast, and bright-field
microscopy. Figure~\ref{fig:dataset_gallery} shows one example from every
source dataset. Thirteen datasets yield materialized benchmark items; the
remaining two contain no anchors after hard-mining selection and completeness
filtering.

Sequences are partitioned before links are ranked or selected. For CTC
datasets, sequence \texttt{01} is assigned to validation and sequence
\texttt{02} to test. Bacterial sequences are ordered by identifier and divided
into lower and upper halves; a two-sequence collection therefore assigns its
lower identifier to validation and its higher identifier to test. The
multi-sequence assignments are \texttt{01--02}/\texttt{03--05} for
\texttt{metA}, \texttt{01}/\texttt{02,07} for \texttt{rpsM}, and
\texttt{03--04}/\texttt{05--06} for \texttt{trpL}, where the two sides of
each slash denote validation and test. The available \texttt{pheA} data
contains only sequence \texttt{05} and is therefore test-only.

\begin{table}[H]
    \centering
    \scriptsize
    \setlength{\tabcolsep}{2.4pt}
    \caption{Materialized benchmark size. Every neighborhood contains exactly
    five source cells; targets are the next-frame candidates for those
    sources. A source with two daughters contributes two division edges.}
    \label{tab:split_stats}
    \begin{tabular}{llrrrrrr}
        \toprule
        Scale & Split & Items & Src. & Tgt. & Edges & Div. & Cont. \\
        \midrule
        hard & validation & 205 & 1{,}025 & 2{,}516 & 1{,}188 & 358 & 830 \\
        hard & test & 244 & 1{,}220 & 3{,}200 & 1{,}426 & 424 & 1{,}002 \\
        full & validation & 626 & 3{,}130 & 8{,}286 & 3{,}779 & 1{,}350 & 2{,}429 \\
        full & test & 911 & 4{,}555 & 12{,}003 & 5{,}262 & 1{,}443 & 3{,}819 \\
        \bottomrule
    \end{tabular}
\end{table}

The full benchmark contains 1,537 neighborhoods and 9,041 annotated linking
decisions. The hard partitions are the subsets used for harness selection and
the frozen leaderboard.

\subsection{Difficulty-Probe Roles and Held-Out Predictions}

\begin{table*}[t]
    \centering
    \small
    \setlength{\tabcolsep}{5pt}
    \caption{The three specialized trackers expose complementary association
    failure modes. Their predictions are used to identify difficult links,
    not as estimates of performance on the unfiltered source datasets.}
    \label{tab:proxy_trackers}
    \resizebox{\textwidth}{!}{%
    \begin{tabular}{llll}
        \toprule
        Probe & Reasoning unit & Information used & Role in construction \\
        \midrule
        Distance Hungarian &
        Detection centroid &
        Pairwise displacement &
        Geometric one-to-one baseline; no divisions \\
        Trackastra &
        Segmented cell &
        Learned appearance, motion, and node context &
        Node-level learned correspondence \\
        HOCT &
        Cell and candidate link &
        Node context and link-to-link interactions &
        High-order correspondence and division reasoning \\
        \bottomrule
    \end{tabular}%
    }
\end{table*}

We divide the 15 source datasets into three folds. For each fold, the learned
trackers are trained on the other two folds and evaluated on the held-out
datasets. Consequently, every prediction used by the mining procedure is
out-of-domain with respect to tracker training. All three probes receive the
provided segmentation masks, so their errors isolate association rather than
detection quality. The distance matcher contributes to the hard-mining count
$q(e)$ but not to the learned-score hardness $h(e)$ because its assignment
cost is not a calibrated link probability.

\subsection{Expected Shared Errors under Independence}
\label{app:shared-error-expectation}

The $0.0022\%$ expectation in Section~\ref{ssec:difficulty_probes} is computed
by stratifying the candidate links by sequence. For each sequence $s$, if
$N_s$ is its number of candidate links and $p_{i,s}$ is tracker $i$'s marginal
error rate, independence predicts
$N_s\prod_{i=1}^{3}p_{i,s}$ shared errors. Summing this quantity across
sequences gives 33.2 expected shared errors, or $0.0022\%$ of the roughly
1.5M links. In contrast, multiplying the trackers' global marginal error rates
of $1.23\%$, $2.93\%$, and $1.33\%$ gives only
$1.5{\times}10^6(0.0123)(0.0293)(0.0133)\approx 7.2$ expected links.
Thus, the per-sequence estimate already allows errors to concentrate in
uniformly harder sequences, yet the 4,222 observed shared errors remain
approximately $127{\times}$ the stratified independence expectation.

\subsection{The Frozen c1 Evaluation Prompt}
\label{app:c1_prompt}

The complete cell-linking prompt below is for one validation item under the
frozen c1 interface. The accompanying image is an unannotated two-panel
overview, and coordinates are normalized independently within each panel.
Track repair and inspection share the header, candidate list, and evidence
block but use task-specific instructions and answer schemas.

{\scriptsize
\begin{verbatim}
Output budget: you have 8192 output tokens in total for this
  item, counting your reasoning and every tool-calling turn,
  not just the final answer. Spend them so that the JSON
  answer still fits. You will be told when 20%

Two consecutive microscopy frames from a cell-tracking
  sequence are shown: frame t (left) and frame t+1 (right).
  Frame t contains 5 labelled source cells; frame t+1 contains
  their candidate target cells.

Candidate targets allowed for each source:
- S1: T1, T2, T3, T4, T5, T6
- S2: T1, T2, T3, T4, T5, T6
- S3: T1, T2, T3, T4, T5, T6
- S4: T1, T2, T3, T4, T5, T6
- S5: T1, T2, T3, T4, T5, T6

A continuation joins a source to the same track in the next
  frame. A division edge joins an annotated parent to each
  visible daughter track; adjacent-frame annotations can
  contain one or more visible daughters, although two is the
  usual case.

Return the complete ground-truth linking for all 5 sources. A
  source may disappear (zero targets), continue (one target),
  or branch to one or more visible daughter targets. Use
  semantic=division for every daughter edge in a branch. The
  number of visible daughters is defined by the lineage
  annotation; do not assume it is exactly two. Do not omit
  links simply because they are not the most obvious error in
  the image.

The image is unannotated. Each cell's location is given below
  as a normalized bounding box [x0, y0, x1, y1], where 0.0-1.0
  is relative to that cell's own panel (the left panel for S
  ids, the right panel for T ids):
  S1: [0.436, 0.582, 0.491, 0.661]
  S2: [0.398, 0.206, 0.435, 0.243]
  S3: [0.449, 0.756, 0.487, 0.794]
  S4: [0.398, 0.483, 0.435, 0.521]
  S5: [0.522, 0.715, 0.559, 0.751]
  T1: [0.398, 0.210, 0.435, 0.247]
  T2: [0.398, 0.483, 0.435, 0.521]
  T3: [0.458, 0.586, 0.495, 0.624]
  T4: [0.436, 0.646, 0.474, 0.684]
  T5: [0.500, 0.697, 0.538, 0.734]
  T6: [0.453, 0.745, 0.491, 0.781]
You may call crop_and_zoom(panel, bbox, zoom) to inspect any
  region more closely, as many times as you need. zoom is
  1.0-4.0.

Think it through, then output only the JSON object, with no
  text after it.
\end{verbatim}
\begin{lstlisting}[style=json]
{"links": [{"source": "S1", "target": "T2", "semantic":
  "continuation"}], "unlinked_sources": ["S4"], "confidence":
  0.0, "reason": "short visual justification"}
\end{lstlisting}
}

\subsubsection{Tool Interface}
\label{app:tool_schema}

Condition c1 exposes one function through each model's native tool-calling
interface. The harness executes valid calls with a deterministic renderer and
returns the crop as a new image turn.

\noindent
\begin{minipage}{\columnwidth}
\begin{lstlisting}[style=json]
[
  {
    "type": "function",
    "name": "crop_and_zoom",
    "description": "Return a magnified crop of a region of one panel.",
    "parameters": {
      "type": "object",
      "properties": {
        "panel": {
          "type": "string",
          "enum": ["t", "t1"],
          "description": "'t' = left/frame t, 't1' = right/frame t+1"
        },
        "bbox": {
          "type": "array",
          "items": {"type": "number"},
          "minItems": 4,
          "maxItems": 4,
          "description": "[x0, y0, x1, y1] normalized 0.0-1.0 within that panel"
        },
        "zoom": {
          "type": "number",
          "description": "magnification, 1.0-4.0"
        }
      },
      "required": ["panel", "bbox", "zoom"],
      "additionalProperties": false
    }
  }
]
\end{lstlisting}
\end{minipage}

\subsection{Harness Selection and Execution}
\label{app:model-harness}

\begin{figure}[t]
  \centering
  \includegraphics[width=\columnwidth]{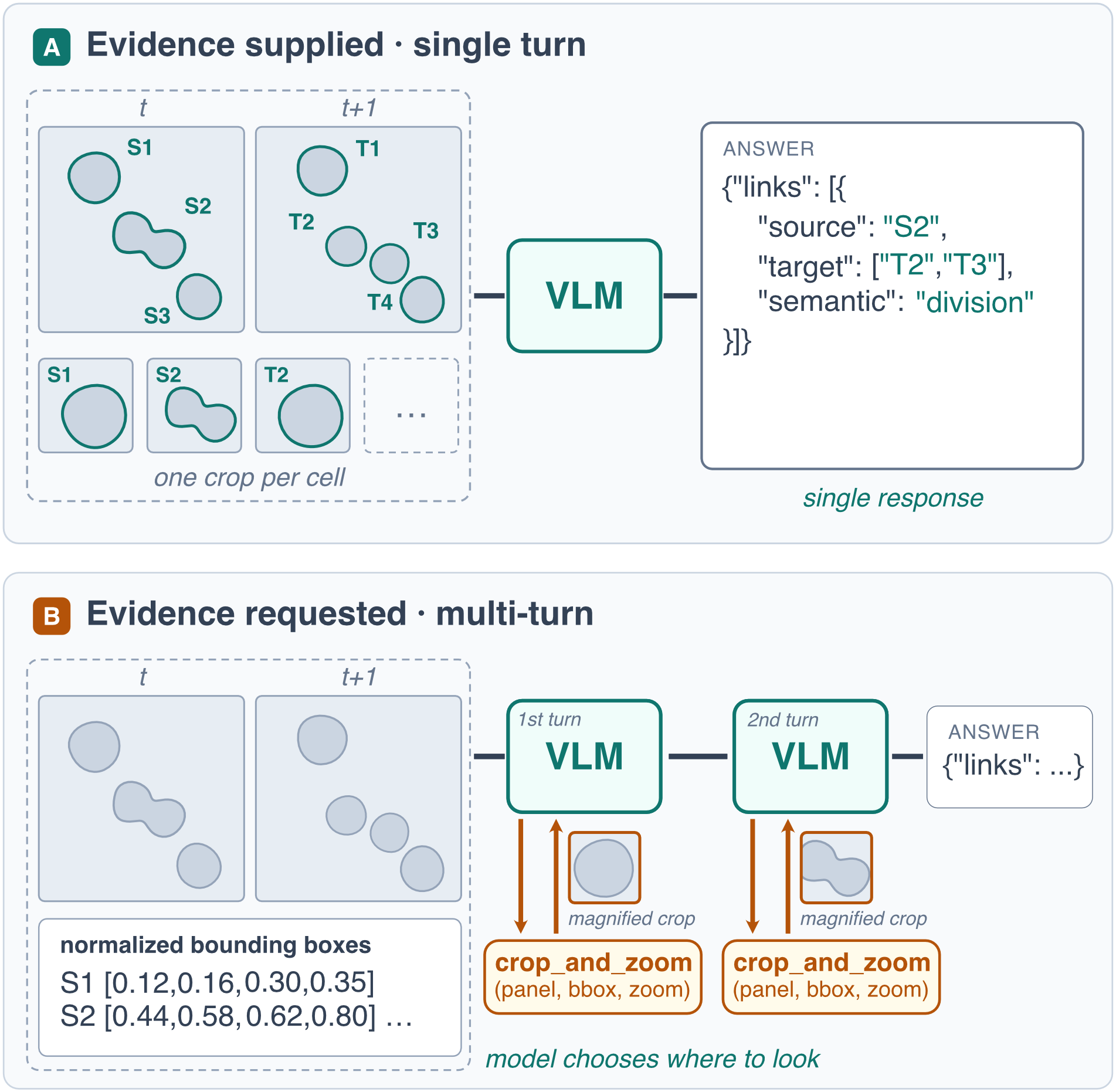}
  \caption{%
  Two evidence-interface families. In the evidence-supplied single-turn
  interface (top), annotated frames and per-cell crops are included in the
  prompt before one response. In the evidence-requested multi-turn interface
  (bottom), the model receives an unannotated overview and normalized bounding
  boxes, then chooses regions to inspect through \texttt{crop\_and\_zoom}
  before returning its answer.%
  }
  \label{fig:multiturn}
\end{figure}

\begin{figure*}[!t]
  \centering
  {%
  \setlength{\tabcolsep}{4pt}%
  \renewcommand{\arraystretch}{1.1}%
  \begin{tabular}{|p{0.47\textwidth}|p{0.47\textwidth}|}
  \hline
  \begin{subfigure}[t]{\linewidth}
    \centering
    \begin{minipage}[b][0.33\linewidth][t]{\linewidth}
    \raggedright
    \includegraphics[width=0.64\linewidth]{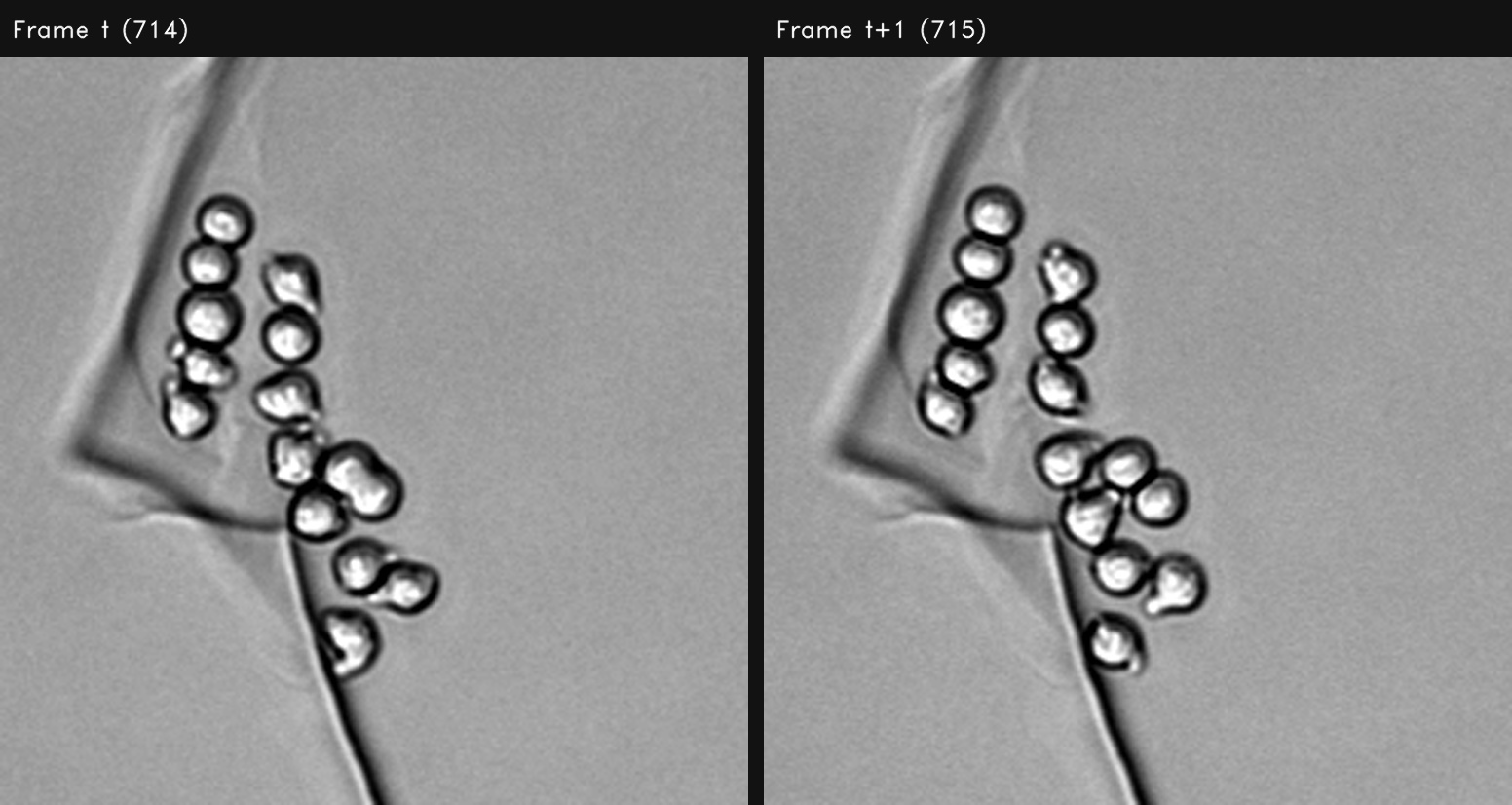}\hspace{3pt}%
    \includegraphics[width=0.295\linewidth]{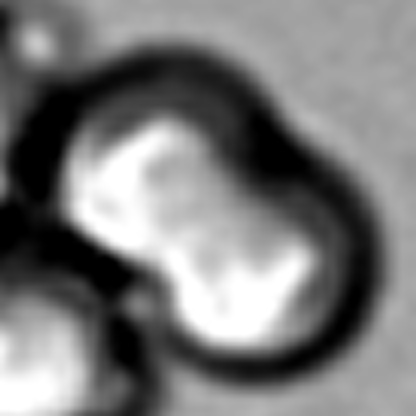}
    \end{minipage}
    \caption{\textbf{c1 --- coordinate baseline.} Unannotated overview is the only image;
    every cell is described by a normalised text bounding box; region tool
    \texttt{crop\_and\_zoom(panel, bbox, zoom)}. Right: example tool response
    (S2 neighbourhood, zoom 4).}
  \end{subfigure} &
  \begin{subfigure}[t]{\linewidth}
    \centering
    \begin{minipage}[b][0.33\linewidth][t]{\linewidth}
    \raggedright
    \includegraphics[width=0.64\linewidth]{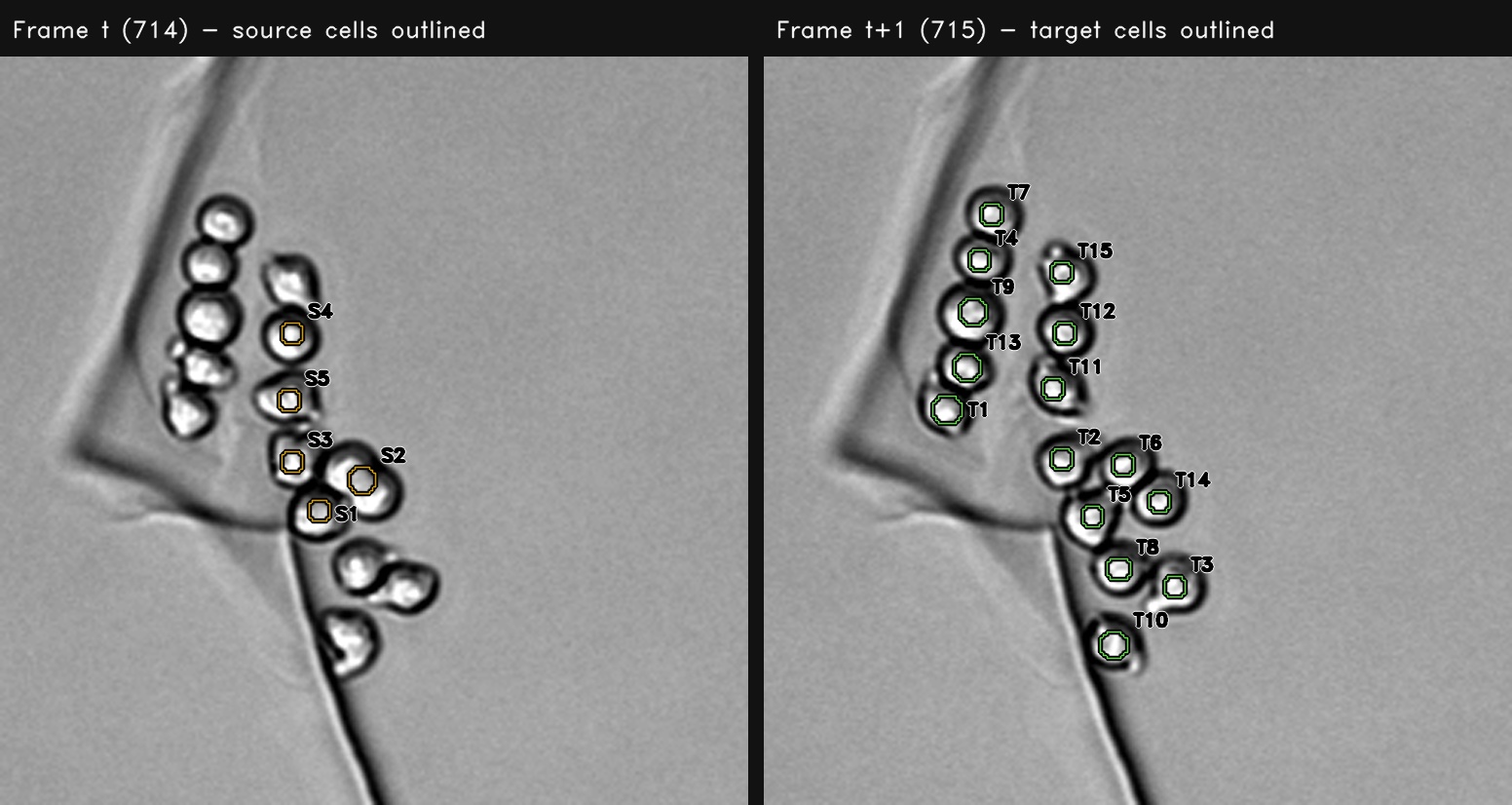}
    \end{minipage}
    \caption{\textbf{c2 --- visual localization.} Sources (amber) and targets (green) outlined
    and labelled in the image; no text coordinates anywhere; same region tool as c1.}
  \end{subfigure}\\ \hline
  \begin{subfigure}[t]{\linewidth}
    \centering
    \begin{minipage}[b][0.50\linewidth][t]{\linewidth}
    \raggedright
    \vspace*{0.17\linewidth}
    \includegraphics[width=0.64\linewidth]{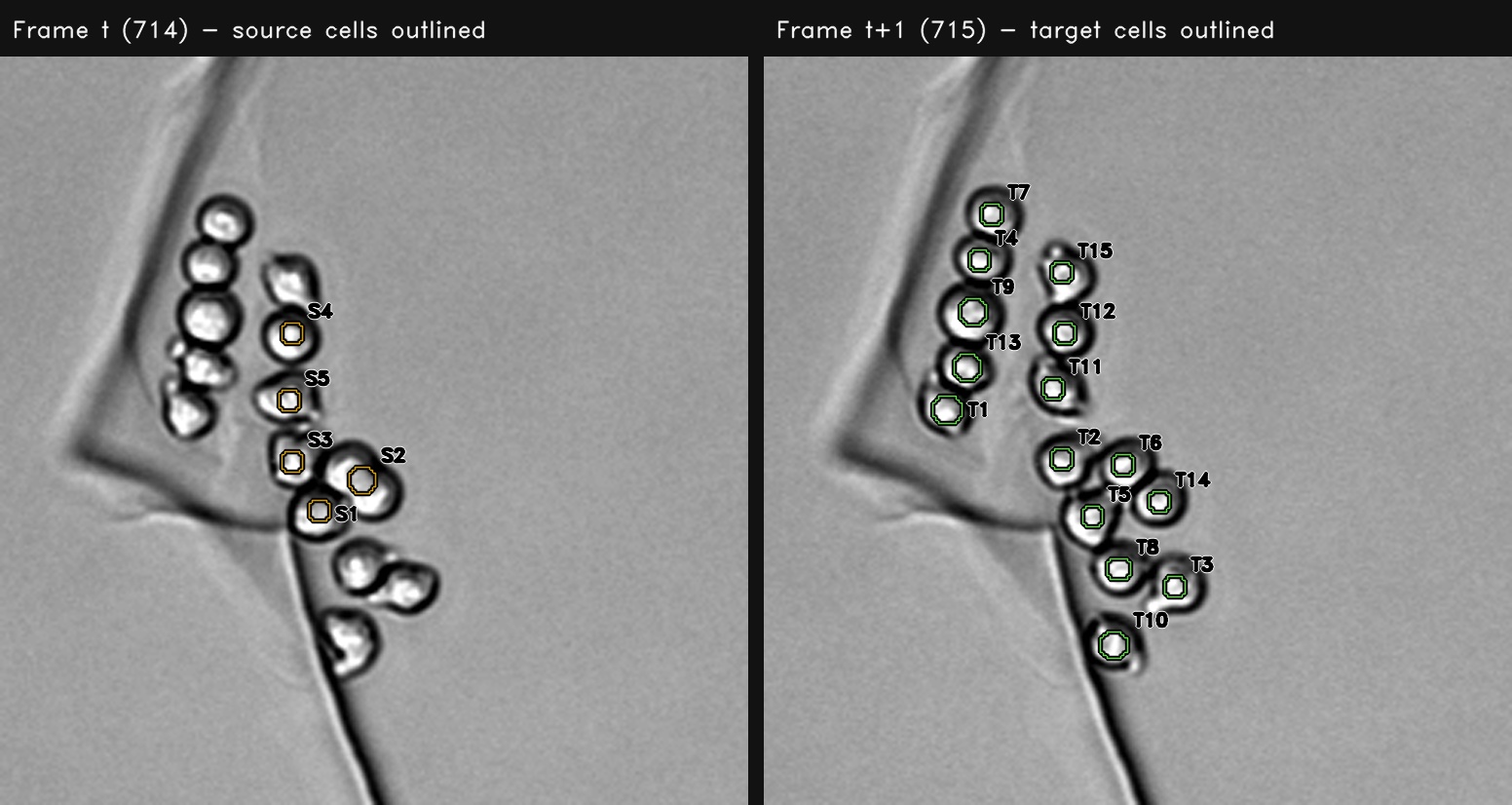}\hspace{3pt}%
    \includegraphics[width=0.295\linewidth]{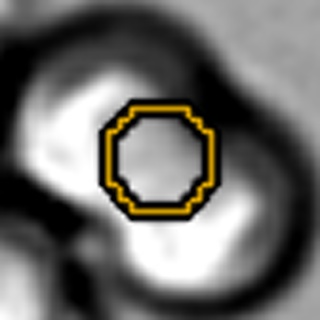}
    \end{minipage}
    \caption{\textbf{c3 --- cell-addressed inspection.} Same annotated overview and text as c2,
    but the closer view is the cell tool \texttt{zoom(cell\_id, zoom)}. Right: example tool
    response to \texttt{zoom("S2", 3.0)}.}
  \end{subfigure} &
  \begin{subfigure}[t]{\linewidth}
    \centering
    \begin{minipage}[b][0.50\linewidth][t]{\linewidth}
    \raggedright
    \includegraphics[width=0.64\linewidth]{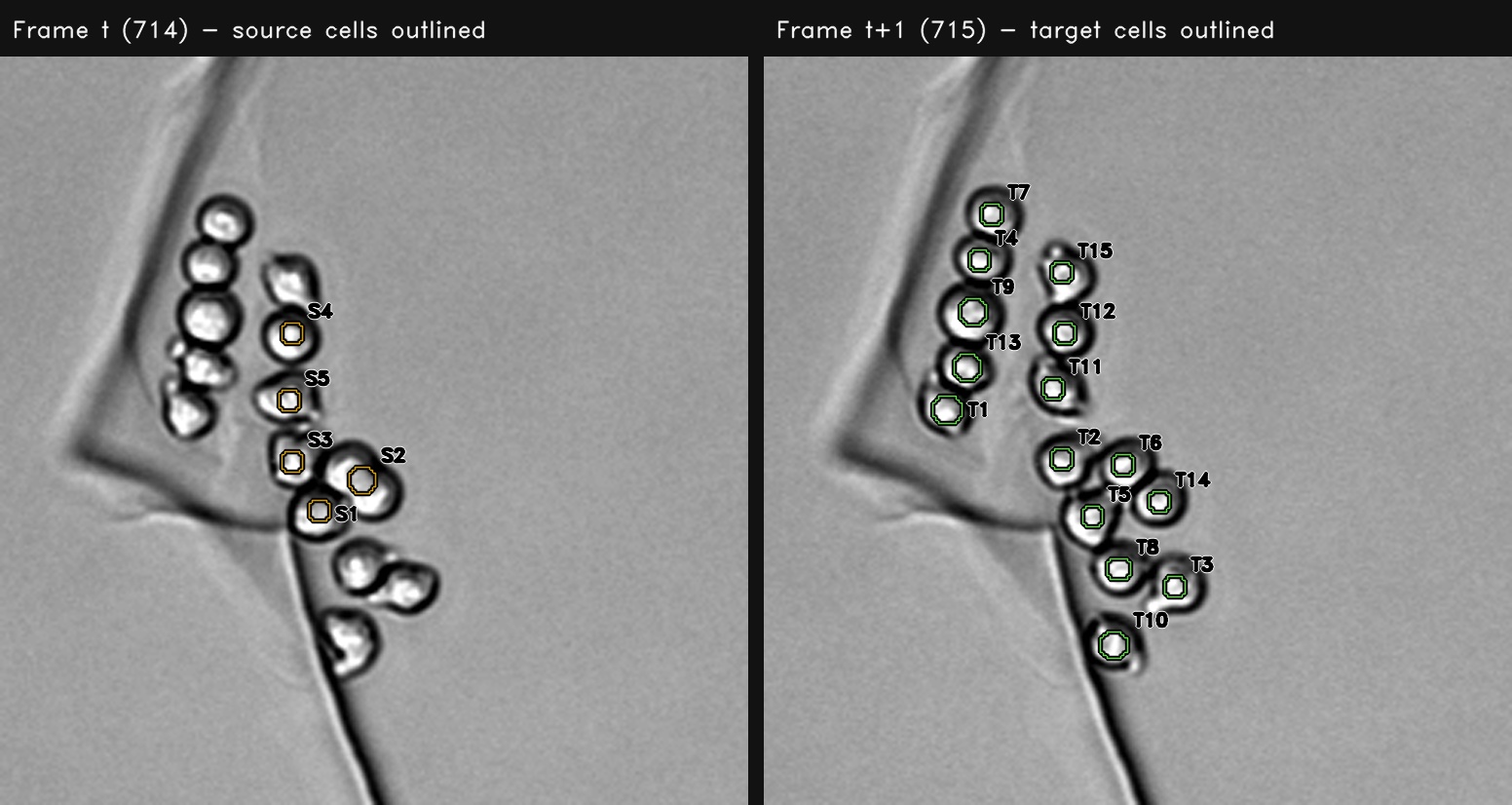}\\[2pt]
    \makebox[0.64\linewidth][c]{%
      \begin{minipage}[c][0.16\linewidth][c]{0.16\linewidth}
        \centering\includegraphics[width=0.94\linewidth]{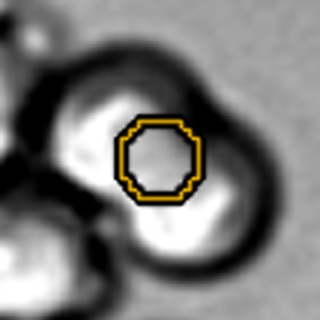}
      \end{minipage}%
      \begin{minipage}[c][0.16\linewidth][c]{0.16\linewidth}
        \centering\includegraphics[width=0.94\linewidth]{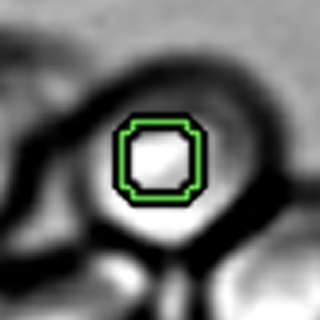}
      \end{minipage}%
      \begin{minipage}[c][0.16\linewidth][c]{0.16\linewidth}
        \centering\includegraphics[width=0.94\linewidth]{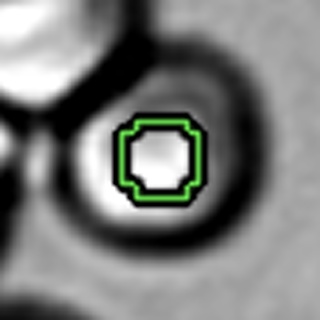}
      \end{minipage}%
      \begin{minipage}[c][0.16\linewidth][c]{0.16\linewidth}
        \centering\scriptsize$+17$\\more crops
      \end{minipage}
    }
    \end{minipage}
    \caption{\textbf{c4 --- fixed visual inspection.} Annotated overview plus one outlined
    close-up per cell, all supplied up front as separate images in id order (S2, T6, T14
    shown of 20); no text coordinates, no tool.}
  \end{subfigure}\\ \hline
  \begin{subfigure}[t]{\linewidth}
    \centering
    \begin{minipage}[b][0.50\linewidth][t]{\linewidth}
    \raggedright
    \includegraphics[width=0.64\linewidth]{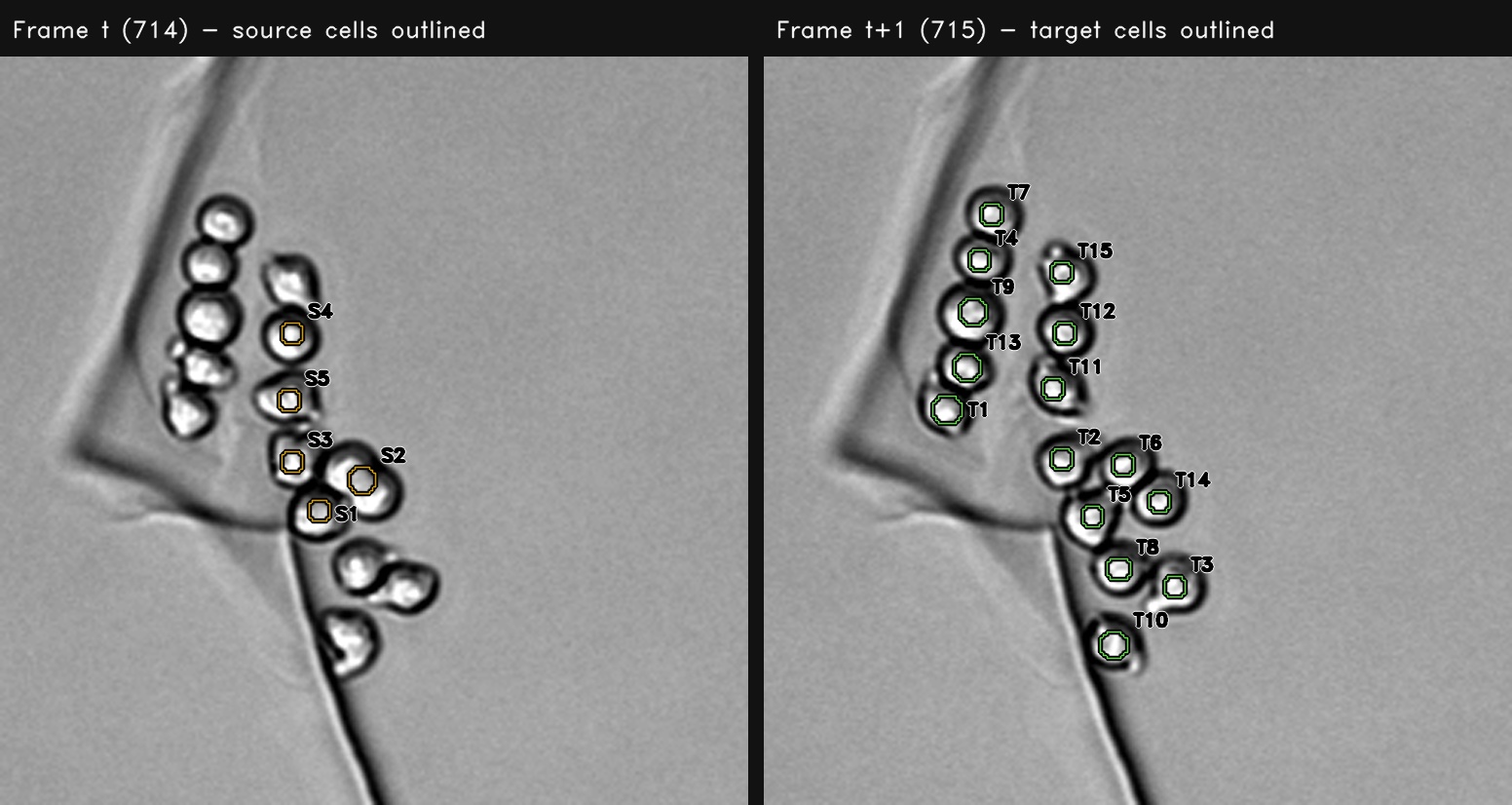}\\[2pt]
    \makebox[0.64\linewidth][c]{%
      \begin{minipage}[c][0.16\linewidth][c]{0.16\linewidth}
        \centering\includegraphics[width=0.94\linewidth]{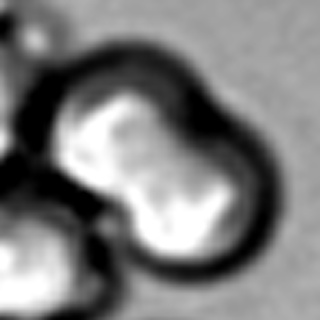}
      \end{minipage}%
      \begin{minipage}[c][0.16\linewidth][c]{0.16\linewidth}
        \centering\includegraphics[width=0.94\linewidth]{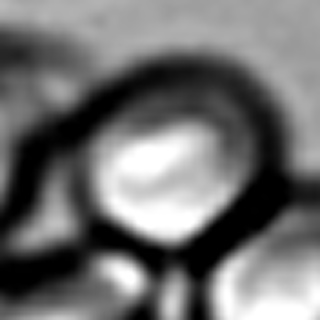}
      \end{minipage}%
      \begin{minipage}[c][0.16\linewidth][c]{0.16\linewidth}
        \centering\includegraphics[width=0.94\linewidth]{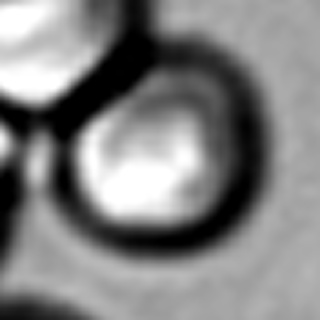}
      \end{minipage}%
      \begin{minipage}[c][0.16\linewidth][c]{0.16\linewidth}
        \centering\scriptsize$+17$\\more crops
      \end{minipage}
    }
    \end{minipage}
    \caption{\textbf{c5 --- crop-outline ablation.} As c4, but the 20 supplied close-ups carry
    no outline; the prompt states the subject is the cell at the centre of each crop. No text
    coordinates, no tool.}
  \end{subfigure} &
  \begin{subfigure}[t]{\linewidth}
    \centering
    \begin{minipage}[b][0.50\linewidth][t]{\linewidth}
    \raggedright
    \vspace*{0.17\linewidth}
    \includegraphics[width=0.64\linewidth]{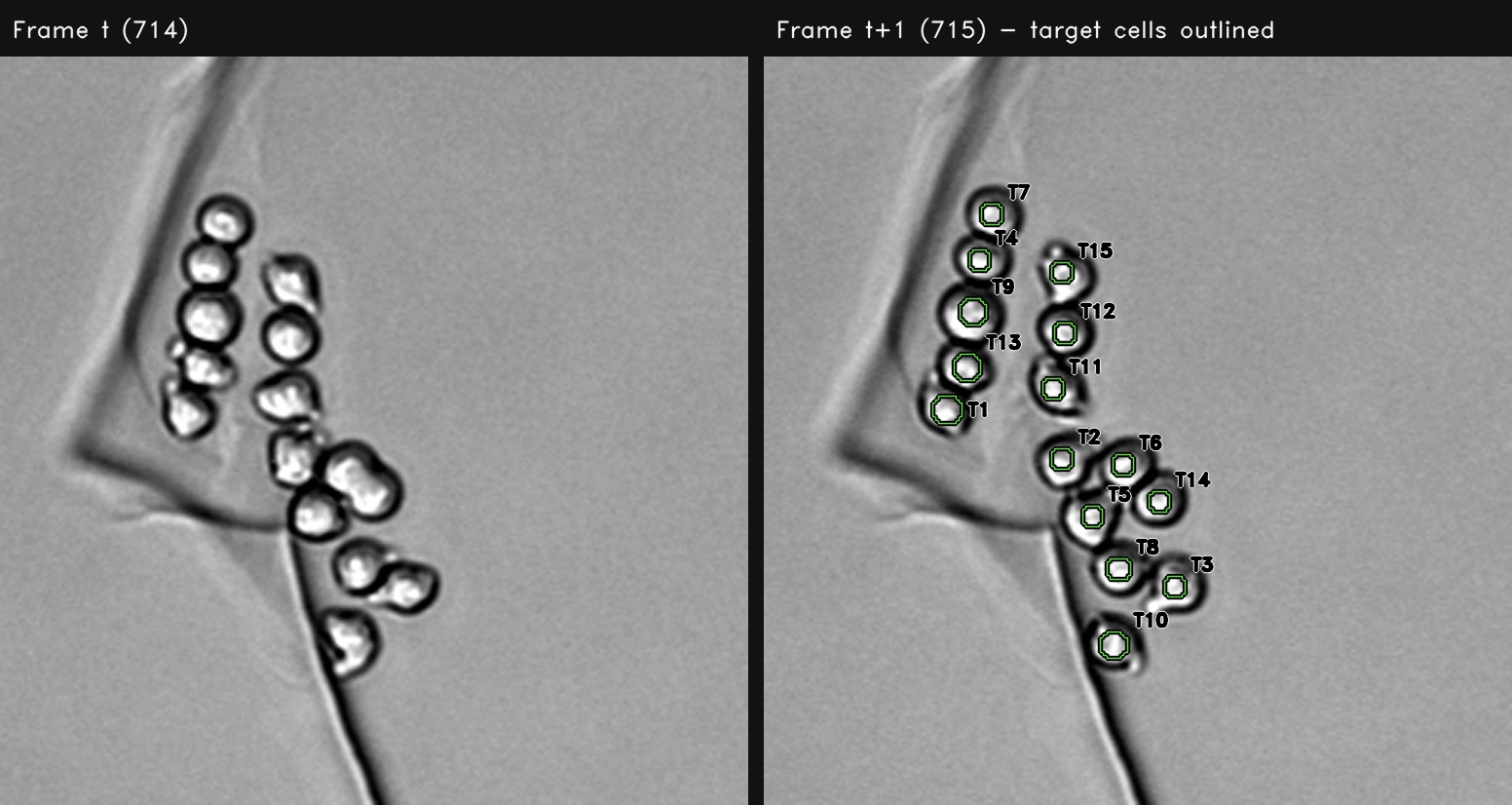}
    \end{minipage}
    \caption{\textbf{c6 --- target-coordinate ablation.} Frame $t$ unannotated with source
    locations given only as text bounding boxes; frame $t{+}1$ targets outlined and labelled
    with no coordinates; same region tool as c1.}
  \end{subfigure}\\ \hline
  \begin{subfigure}[t]{\linewidth}
    \centering
    \begin{minipage}[b][0.33\linewidth][t]{\linewidth}
    \raggedright
    \vspace*{0.17\linewidth}
    \makebox[0.64\linewidth][c]{%
      \begin{minipage}[c][0.16\linewidth][c]{0.16\linewidth}
        \centering\includegraphics[width=0.94\linewidth]{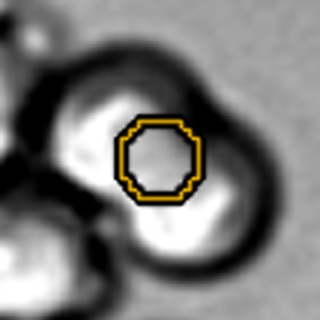}
      \end{minipage}%
      \begin{minipage}[c][0.16\linewidth][c]{0.16\linewidth}
        \centering\includegraphics[width=0.94\linewidth]{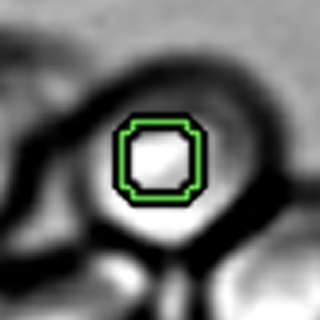}
      \end{minipage}%
      \begin{minipage}[c][0.16\linewidth][c]{0.16\linewidth}
        \centering\includegraphics[width=0.94\linewidth]{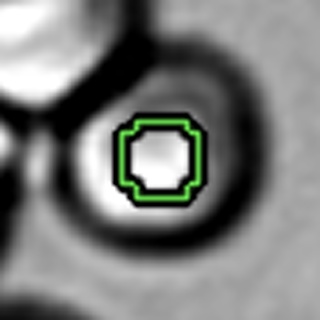}
      \end{minipage}%
      \begin{minipage}[c][0.16\linewidth][c]{0.16\linewidth}
        \centering\scriptsize$+17$\\more crops
      \end{minipage}
    }
    \end{minipage}
    \caption{\textbf{c7 --- global-position ablation.} No overview at all: the 20 outlined
    close-ups (identical renders to c4's), sent in id order, are the only images; no text
    coordinates, no tool. Diagnostic only, not a deployable interface.}
  \end{subfigure} &
  \begin{subfigure}[t]{\linewidth}
    \centering
    \begin{minipage}[b][0.33\linewidth][t]{\linewidth}
    \raggedright
    \includegraphics[width=0.64\linewidth]{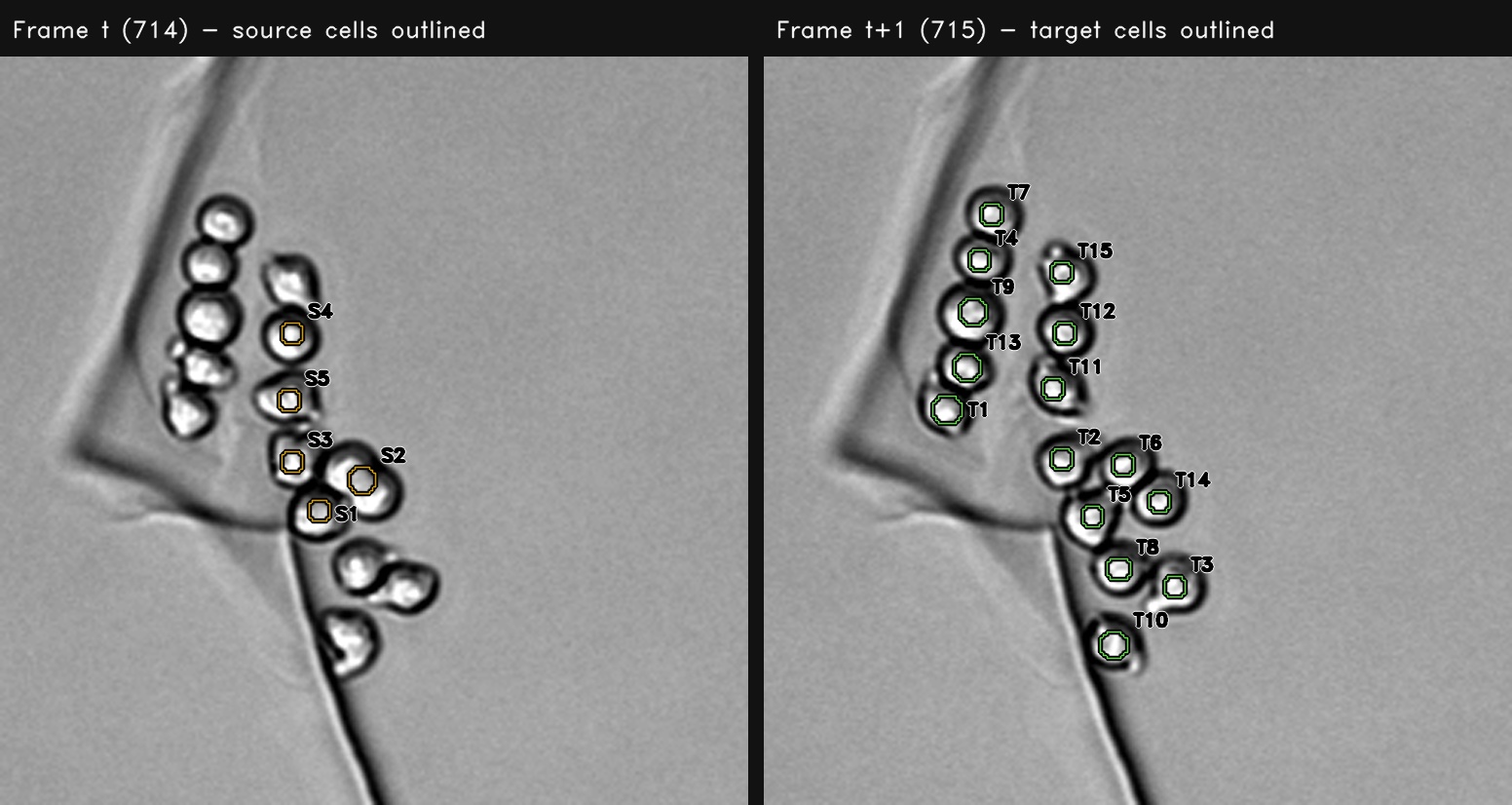}
    \end{minipage}
    \caption{\textbf{c8 --- combined evidence.} Text bounding boxes for every cell (as in c1)
    \emph{and} the same cells outlined and labelled in the image (as in c2), so each cell is
    identified twice; same region tool as c1.}
  \end{subfigure}\\ \hline
  \end{tabular}%
  }
  \caption{All eight evidence-harness conditions (definitions in
  Table~\ref{tab:evidence_harnesses}), instantiated on one fixed division-bearing test item
  (\texttt{BF-C2DL-HSC}, sequence \texttt{02}, frames 714--715: five source cells and fifteen
  candidate targets; parent S2 divides into daughters T6 and T14). Every panel is the exact
  image the harness serves for the cell-linking task. The task text, candidate lists, and
  answer schema remain fixed, so the conditions differ only in the evidence shown.}
  \label{fig:conditions}
\end{figure*}

\begin{table}[t]
    \centering
    \small
    \setlength{\tabcolsep}{5pt}
    \caption{Validation-only harness selection with \texttt{gpt-5.5} at low
    reasoning effort. Entries are dataset-macro primary scores; Mean is their
    unweighted average across the three tasks. Condition c7 is a diagnostic
    information-loss ablation.}
    \label{tab:harness_validation}
    \begin{tabular}{lrrrr}
        \toprule
        Condition & Linking & Repair & Inspection & Mean \\
        \midrule
        \textbf{c1} & \textbf{0.662} & \textbf{0.690} & \textbf{0.474} &
        \textbf{0.609} \\
        c2 & 0.423 & 0.535 & 0.261 & 0.406 \\
        c3 & 0.524 & 0.639 & 0.367 & 0.510 \\
        c4 & 0.444 & 0.546 & 0.348 & 0.446 \\
        c5 & 0.447 & 0.544 & 0.296 & 0.429 \\
        c6 & 0.375 & 0.479 & 0.283 & 0.379 \\
        c7$^{\dagger}$ & 0.199 & 0.330 & 0.189 & 0.239 \\
        c8 & 0.592 & 0.641 & 0.409 & 0.547 \\
        \bottomrule
    \end{tabular}
\end{table}

Table~\ref{tab:harness_validation} reports the primary-score statistic used
to select c1; Table~\ref{tab:evidence_harnesses} additionally reports
NN-residual linking F1. The complete validation sweep contains 4,920
responses: 205 neighborhoods, three tasks, and eight evidence conditions.

Models are invoked through their native multimodal interface and, when tools
are available, their native tool-call representation. The harness maps tool
calls to the same deterministic server-side crop operation and normalizes only
the final answer to the common task representation used for scoring. We draw
one completion for every instance, task, and condition. Parallel dispatch can
change execution order but not prompts, predictions, or scores.

Before a complete sweep, a fixed ten-instance smoke set spanning datasets
checks prompt completeness, image rendering, native tool execution, final
answer parsing, and scorer coverage for every task and condition. A complete
sweep is invalidated by missing calls, terminal-answer truncation,
output-budget exhaustion, tool-loop failure, or provenance mismatch. API
request failures are retried at most five times under a fixed policy and never
according to answer quality.

Each request receives 8,192 output tokens shared by reasoning and tool turns,
with 2,048 tokens reserved for the final structured answer. If an inspection
turn reaches its cap, tools are withdrawn and the reserved answer turn is
invoked. Individual malformed answers remain model outcomes and receive the
score implied by their parsed or empty prediction; JSON and schema failures are
reported per condition. The selection sweep produced no schema failures,
terminal truncations, budget exhaustions, safety-cap hits, or tool failures.

\subsection{Metric Detail}
\label{app:evaluation}

The primary score for linking is F1 over
$(\mathrm{source},\mathrm{target},\mathrm{semantic})$ edge tuples. Repair
uses the same score after applying the predicted operations to the HOCT
proposal, and inspection uses F1 over review decisions. NN-residual F1 removes
links explained by nearest-centroid matching. Division F1 applies the semantic
edge score to division links, exposing performance on the infrequent branch
class without replacing the primary score.

\subsubsection{Subgraph-LNK}
\label{app:subgraph_lnk}

Subgraph-LNK adapts the AOGM/CTC linking score~\cite{matula2015aogm} to the
benchmark's local neighborhoods. For item $i$, the ground-truth and predicted
edge sets contain tuples $(\mathrm{source},\mathrm{target},\mathrm{semantic})$,
where the semantic label is continuation or division. The edit cost is
\[
C_A^{(i)}
=1.5N_{\mathrm{miss}}^{(i)}
+1.0N_{\mathrm{spur}}^{(i)}
+1.0N_{\mathrm{sem}}^{(i)},
\]
where the terms count missing endpoint pairs, spurious endpoint pairs, and
matching endpoint pairs with the wrong semantic label. A semantic error is
charged once rather than as one deletion and one addition.

\begin{table}[t]
    \centering
    \scriptsize
    \setlength{\tabcolsep}{3pt}
    \caption{Edit operations used by Subgraph-LNK.}
    \label{tab:aogm_costs}
    \begin{tabular}{lp{4.5cm}r}
        \toprule
        Operation & Charged when & Cost \\
        \midrule
        Addition & A ground-truth endpoint pair is missing & 1.5 \\
        Deletion & A predicted endpoint pair is spurious & 1.0 \\
        Semantic change & Endpoints match but the edge label is wrong & 1.0 \\
        \bottomrule
    \end{tabular}
\end{table}

At benchmark level, neighborhoods are treated as the disjoint union of one
graph:
\[
\begin{aligned}
\mathrm{LNK}_{\mathrm{sub}}
    &=1-\frac{\min(C_A,C_0)}{C_0},\\
C_A&=\sum_i C_A^{(i)},&
C_0&=1.5\sum_i |E_{\mathrm{gt}}^{(i)}|.
\end{aligned}
\]
The cap ensures that a prediction requiring more edits than construction from
an empty graph receives zero rather than a negative score. Perfect and empty
predictions therefore score one and zero, respectively. On the hard test
split, $C_0=2{,}139.0$ for 1,426 ground-truth edges.

For linking, the predicted graph is scored directly and $\Delta$NN denotes its
margin over the nearest-centroid graph. For repair, predicted add, delete, and
semantic-change operations are applied to the HOCT proposal before scoring;
$\Delta$LNK is measured against the unedited proposal. For inspection, every
flagged source receives a simulated oracle correction: all proposal out-edges
from that source are replaced by its ground-truth out-edges. We report both
the resulting $\Delta$LNK and the gain per 1,000 flagged edges. Missing or
unparseable linking answers are treated as empty predictions; missing repair
or inspection answers leave the proposal unchanged.

This local score preserves the graph-edit interpretation of CTC LNK while
holding segmentation fixed. It is therefore comparable in spirit, but not
numerically identical, to full-sequence LNK. We do not report the CTC composite
OP\textsubscript{CLB}: because segmentation is supplied, its detection
component would be degenerate.

\onecolumn
\section{Additional Quantitative Results}
\label{app:additional-results}

\begin{table}[H]
    \centering
    \scriptsize
    \setlength{\tabcolsep}{4pt}
    \renewcommand{\arraystretch}{0.95}
    \caption{Validation--test robustness for GPT-5.6 (terra). Validation
    contains 205 neighborhoods and test contains 244. NN, HOCT, and Terra
    report LNK$_{\mathrm{sub}}$; Mean is the unweighted mean of the three
    task-primary scores. Repair $\Delta$LNK is measured relative to the
    split-specific HOCT proposal. Inspection @10 and Gold @10 use the tightest
    review budget. Validation results are descriptive because that split was
    used to select the harness and contains one system and completion.}
    \label{tab:validation_robustness}
    \resizebox{\textwidth}{!}{%
    \begin{tabular}{lrrrrrrrrr}
        \toprule
        Split
        & NN LNK
        & HOCT LNK
        & Terra LNK
        & $\Delta$NN
        & Mean
        & Repair F1
        & Repair $\Delta$LNK
        & Inspection @10
        & Gold @10 \\
        \midrule
        Validation
            & 0.409
            & 0.527
            & 0.616
            & $+0.207$
            & 0.730
            & 0.791
            & $+0.154$
            & $+0.210$
            & $+0.268$ \\
        Test
            & 0.493
            & 0.385
            & 0.692
            & $+0.199$
            & 0.724
            & 0.788
            & $+0.337$
            & $+0.244$
            & $+0.284$ \\
        \bottomrule
    \end{tabular}%
    }
\end{table}

\begin{table}[H]
    \centering
    \scriptsize
    \setlength{\tabcolsep}{4pt}
    \renewcommand{\arraystretch}{0.93}
    \caption{Track-repair decomposition, sorted by graph-level benefit.
    $\Delta$LNK is measured relative to the HOCT proposal; the oracle ceiling
    is $+0.615$. Add and Delete report micro-averaged precision/recall against
    the gold edit set. Better and Worse denote the percentage of neighborhoods
    whose per-item LNK increases or decreases; the remainder are unchanged.}
    \label{tab:repair_breakdown}
    \resizebox{\textwidth}{!}{%
    \begin{tabular}{lrrccrr}
        \toprule
        System
        & Repair F1
        & $\Delta$LNK
        & Add P/R
        & Delete P/R
        & Better
        & Worse \\
        \midrule
        GPT-5.6 (sol)
            & 0.815 & $+0.412$ & 0.85/0.77 & 0.90/0.77 & 80\% & 7\% \\
        Claude Fable 5
            & 0.786 & $+0.348$ & 0.77/0.72 & 0.82/0.77 & 72\% & 13\% \\
        GPT-5.6 (terra)
            & 0.788 & $+0.337$ & 0.76/0.71 & 0.91/0.67 & 72\% & 12\% \\
        GPT-5.5
            & 0.741 & $+0.280$ & 0.69/0.68 & 0.78/0.73 & 61\% & 15\% \\
        Claude Opus 5
            & 0.739 & $+0.263$ & 0.76/0.58 & 0.78/0.71 & 57\% & 12\% \\
        Claude Opus 4.8
            & 0.729 & $+0.201$ & 0.62/0.58 & 0.77/0.68 & 55\% & 17\% \\
        GPT-5.6 (luna)
            & 0.697 & $+0.180$ & 0.59/0.56 & 0.75/0.53 & 48\% & 28\% \\
        Qwen3.5-397B-A17B
            & 0.656 & $+0.166$ & 0.72/0.37 & 0.85/0.44 & 35\% & 14\% \\
        MiniMax M3
            & 0.656 & $+0.139$ & 0.88/0.26 & 0.95/0.25 & 24\% & 3\% \\
        Kimi K3
            & 0.659 & $+0.111$ & 0.91/0.20 & 0.93/0.19 & 17\% & 2\% \\
        Claude Haiku 4.5
            & 0.629 & $+0.097$ & 0.54/0.45 & 0.75/0.53 & 37\% & 23\% \\
        Qwen3.5-122B-A10B
            & 0.598 & $+0.055$ & 0.58/0.16 & 0.72/0.19 & 16\% & 13\% \\
        Gemma 4 31B
            & 0.606 & $+0.030$ & 0.37/0.46 & 0.50/0.59 & 36\% & 29\% \\
        Claude Sonnet 5
            & 0.631 & $+0.010$ & 0.39/0.09 & 0.48/0.09 & 6\% & 15\% \\
        Qwen3.5-35B-A3B
            & 0.622 & $+0.008$ & 0.47/0.03 & 0.52/0.04 & 2\% & 7\% \\
        InternVL3.5-38B
            & 0.556 & $-0.085$ & 0.13/0.04 & 0.35/0.07 & 3\% & 25\% \\
        GLM-4.6V
            & 0.519 & $-0.094$ & 0.19/0.10 & 0.25/0.16 & 11\% & 57\% \\
        Molmo2-8B
            & 0.521 & $-0.099$ & 0.05/0.03 & 0.06/0.04 & 5\% & 64\% \\
        Mistral Medium 3.5 128B
            & 0.465 & $-0.244$ & 0.12/0.12 & 0.24/0.19 & 6\% & 33\% \\
        Inkling-Small
            & 0.443 & $-0.332$ & 0.20/0.37 & 0.34/0.53 & 14\% & 45\% \\
        Llama 4 Scout
            & 0.342 & $-0.385$ & 0.09/0.24 & 0.27/0.36 & 2\% & 74\% \\
        \bottomrule
    \end{tabular}%
    }
\end{table}

\begin{table}[H]
    \centering
    \scriptsize
    \setlength{\tabcolsep}{4.2pt}
    \renewcommand{\arraystretch}{0.93}
    \caption{Inspection under fixed review budgets. Each entry reports the
    $\Delta$LNK recovered after oracle correction of the model's selected
    sources. Nominal budgets of 10/25/50\% realize 19/38/58\% of proposal
    edges because each neighborhood contains only 4--6 edges. Stored answers
    do not contain confidence scores, so flags are retained in emission order.
    Slash-separated values represent the two systems named in the row; ranges
    are used where only the paired summary was retained.}
    \label{tab:triage_budget}
    \resizebox{\textwidth}{!}{%
    \begin{tabular}{lrrrrrr}
        \toprule
        System
        & Inspection F1
        & Flags
        & @10\%
        & @25\%
        & @50\%
        & All flags \\
        \midrule
        \emph{Gold-flag oracle}
            & -- & 1061
            & $+0.284$ & $+0.441$ & $+0.542$ & $+0.615$ \\
        \midrule
        GPT-5.6 (sol)
            & 0.631 & 932
            & $+0.249$ & $+0.265$ & $+0.387$ & $+0.518$ \\
        Claude Fable 5
            & 0.698 & 993
            & $+0.249$ & $+0.287$ & $+0.396$ & $+0.513$ \\
        GPT-5.6 (terra)
            & 0.590 & 978
            & $+0.244$ & $+0.288$ & $+0.413$ & $+0.518$ \\
        GPT-5.5
            & 0.588 & 979
            & $+0.235$ & $+0.264$ & $+0.378$ & $+0.500$ \\
        Claude Opus 5
            & 0.532 & 847
            & $+0.223$ & $+0.250$ & $+0.356$ & $+0.459$ \\
        Claude Opus 4.8
            & 0.499 & 866
            & $+0.205$ & $+0.227$ & $+0.326$ & $+0.437$ \\
        GPT-5.6 (luna)
            & 0.359 & 919
            & $+0.194$ & $+0.235$ & $+0.350$ & $+0.463$ \\
        Claude Haiku 4.5
            & 0.266 & 543
            & $+0.169$ & $+0.226$ & $+0.278$ & $+0.307$ \\
        Qwen3.5 397B / 122B
            & 0.377/0.227
            & 628/617
            & $+0.153$
            & $+0.18$--$+0.20$
            & $+0.27$
            & $+0.342$/$+0.322$ \\
        Gemma 4 31B
            & 0.267 & 604
            & $+0.145$ & $+0.178$ & $+0.254$ & $+0.314$ \\
        Inkling-Small
            & 0.195 & 1376
            & $+0.141$ & $+0.229$ & $+0.333$ & $+0.521$ \\
        GLM-4.6V
            & 0.183 & 716
            & $+0.138$ & $+0.208$ & $+0.270$ & $+0.318$ \\
        Molmo2-8B
            & 0.015 & 718
            & $+0.120$ & $+0.241$ & $+0.365$ & $+0.367$ \\
        Llama 4 Scout
            & 0.098 & 3061
            & $+0.117$ & $+0.194$ & $+0.243$ & $+0.519$ \\
        MiniMax M3
            & 0.183 & 371
            & $+0.089$ & $+0.124$ & $+0.169$ & $+0.209$ \\
        Claude Sonnet 5
            & 0.079 & 369
            & $+0.068$ & $+0.106$ & $+0.162$ & $+0.174$ \\
        Mistral Medium 3.5 128B
            & 0.084 & 999
            & $+0.059$ & $+0.090$ & $+0.129$ & $+0.216$ \\
        Kimi K3
            & 0.083 & 230
            & $+0.052$ & $+0.074$ & $+0.107$ & $+0.128$ \\
        Qwen3.5 35B / InternVL3.5
            & 0.076/0.026
            & 189/308
            & $+0.04$--$+0.05$
            & $+0.05$--$+0.09$
            & $+0.07$--$+0.12$
            & $+0.094$/$+0.129$ \\
        \midrule
        \emph{Random reviewer}
            & -- & --
            & $+0.125$ & $+0.245$ & $+0.371$ & -- \\
        \bottomrule
    \end{tabular}%
    }
\end{table}

\begin{table}[H]
    \centering
    \scriptsize
    \setlength{\tabcolsep}{3.5pt}
    \renewcommand{\arraystretch}{0.93}
    \caption{Per-neighborhood linking LNK$_{\mathrm{sub}}$ by difficulty
    stratum, macro-averaged over datasets represented within each stratum.
    NN-solved denotes neighborhoods exactly reconstructed by nearest-centroid.
    EA, EC, and ED denote missing, semantic, and spurious anchor errors.}
    \label{tab:difficulty_strata}
    \begin{tabular}{lrrrr}
        \toprule
        Stratum
        & $n$
        & NN baseline
        & GPT-5.6 (sol)
        & Gemma 4 \\
        \midrule
        NN-solved
            & 22 & 1.000 & 0.992 & 0.976 \\
        NN-resistant
            & 222 & 0.494 & 0.766 & 0.536 \\
        \midrule
        Hardness Q1 (easiest)
            & 48 & 0.706 & 0.944 & 0.629 \\
        Hardness Q2
            & 49 & 0.707 & 0.901 & 0.718 \\
        Hardness Q3
            & 49 & 0.432 & 0.689 & 0.411 \\
        Hardness Q4
            & 49 & 0.510 & 0.662 & 0.533 \\
        Hardness Q5 (hardest)
            & 49 & 0.459 & 0.623 & 0.473 \\
        \midrule
        Proxy EA (missing)
            & 206 & 0.500 & 0.766 & 0.524 \\
        Proxy EC (semantic)
            & 31 & 0.472 & 0.735 & 0.550 \\
        Proxy ED (spurious)
            & 7 & 0.748 & 0.906 & 0.664 \\
        \midrule
        Division-bearing
            & 181 & 0.498 & 0.824 & 0.528 \\
        No division
            & 63 & 0.720 & 0.787 & 0.742 \\
        \bottomrule
    \end{tabular}
\end{table}

\end{document}